\documentclass[10pt,twocolumn,letterpaper]{article}

\usepackage[pagenumbers]{wacv}
\usepackage[pagebackref,breaklinks,colorlinks]{hyperref}
\usepackage[dvipsnames]{xcolor}
\usepackage{amsmath}
\usepackage{amssymb}
\usepackage{enumitem}
\usepackage{multirow}
\usepackage{microtype}
\usepackage{pgfplotstable}
\pgfplotsset{compat=1.13}
\renewcommand\paragraph[1]{\noindent \textbf{#1}}

\def\SL2D{\ensuremath S_{\text{\tiny L2D}}}
\def\righttarrow{\hspace{-2pt}\rightarrow\hspace{-2pt}}

\def\l1{\ensuremath{\ell_1}\xspace}
\def\l2{\ensuremath{\ell_2}\xspace}

\newcommand{\red}[1]{{\color{Red}{#1}}}

\newcommand{\green}[1]{{\color{Green}{#1}}}

\def\nsp{\hspace{-3pt}}
\def\xssp{\hspace{1pt}}
\def\xtssp{\hspace{2pt}}
\def\ssp{\hspace{4pt}}
\def\msp{\hspace{5pt}}

\def\mllsp{\hspace{10pt}}
\def\lsp{\hspace{12pt}}
\def\xlsp{\hspace{20pt}}

\makeatletter
\DeclareRobustCommand\onedot{\futurelet\@let@token\@onedot}
\def\@onedot{\ifx\@let@token.\else.\null\fi\xspace}
 
\def\ie{\emph{i.e}\onedot}

\def\etal{\emph{et al}\onedot}
\makeatother

\begin{document}

\title{Crafting Distribution Shifts for Validation and Training \\ in Single Source Domain Generalization}
\author{Nikos Efthymiadis \hspace{9ex} Giorgos Tolias \hspace{11ex} Ondřej Chum \\
VRG, Faculty of Electrical Engineering, Czech Technical University in Prague\\
{\tt\small \hspace{3ex} efthynik@fel.cvut.cz \hspace{5ex}  toliageo@fel.cvut.cz \hspace{3ex} chum@cmp.felk.cvut.cz}
}
\maketitle

\begin{abstract}
Single-source domain generalization attempts to learn a model on a source domain and deploy it to unseen target domains. Limiting access only to source domain data imposes two key challenges -- how to train a model that can generalize and how to verify that it does.
The standard practice of validation on the training distribution does not accurately reflect the model's generalization ability, while validation on the test distribution is a malpractice to avoid.
In this work, we construct an independent validation set by transforming source domain images with a comprehensive list of augmentations, covering a broad spectrum of potential distribution shifts in target domains. We demonstrate a high correlation between validation and test performance for multiple methods and across various datasets. The proposed validation achieves a relative accuracy improvement over the standard validation equal to 15.4\% or 1.6\% when used for method selection or learning rate tuning, respectively.

Furthermore, we introduce a novel family of methods that increase the shape bias through enhanced edge maps.
To benefit from the augmentations during training and preserve the independence of the validation set, a $k$-fold validation process is designed to separate the augmentation types used in training and validation.
The method that achieves the best performance on the augmented validation is selected from the proposed family. It achieves state-of-the-art performance on various standard benchmarks. Code at: \url{https://github.com/NikosEfth/crafting-shifts}

\end{abstract}

\vspace{-10pt}

\section{Introduction}
\label{sec:intro}

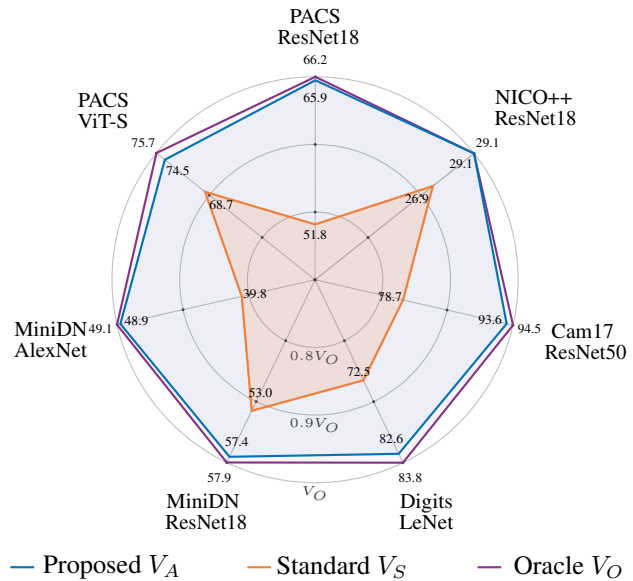
\begin{figure}[t]
\centering
\def\n{7}   
\def\N{3}   
\def\Size{2}
\def\minval{0.7}
\def\rotate{38.5}
\begin{tikzpicture}[scale=0.45]

\newcommand{\calculatefx}[2]{
    \pgfmathsetmacro{\result}{\N*\Size*(#2-\minval)/(1-\minval)}
    \global\expandafter\let\csname#1\endcsname\result
}
\calculatefx{ax0}{0.8}
\calculatefx{ax1}{0.9}
\calculatefx{ax2}{1.0}
\draw[thin, lightgray] (0,0) circle [radius=\csname ax0\endcsname];
\draw(0,-\csname ax0\endcsname -0.3)node[text width=1cm,align=center, color=darkgray]{\tiny $0.8V_O$}; 
\draw[thin, lightgray] (0,0) circle [radius=\csname ax1\endcsname];
\draw(0,-\csname ax1\endcsname -0.3)node[text width=1cm,align=center, color=darkgray]{\tiny $0.9V_O$}; 
\draw[thin, lightgray] (0,0) circle [radius=\csname ax2\endcsname];
\draw(0,-\csname ax2\endcsname -0.3)node[text width=1cm,align=center, color=darkgray]{\tiny $V_O$}; 

\foreach \x in{0,1,...,\n}{%
    \draw[thin,lightgray] (0,0)--(\rotate+360/\n*\x:\Size*\N+0);
    \foreach \y in{0,1,...,\N}{
        \draw[fill=lightgray, darkgray] (\rotate+360/\n*\x:\y*\Size) circle(0.75pt);
        }
}

\calculatefx{fx1}{0.78184}
\calculatefx{fx2}{0.90803}
\calculatefx{fx3}{0.81100}
\calculatefx{fx4}{0.91518}
\calculatefx{fx5}{0.86496}
\calculatefx{fx6}{0.83339}
\calculatefx{fx7}{0.92239}

\calculatefx{fx8}{0.99486}
\calculatefx{fx9}{0.98414}
\calculatefx{fx10}{0.99491}
\calculatefx{fx11}{0.99067}
\calculatefx{fx12}{0.98545}
\calculatefx{fx13}{0.99037}
\calculatefx{fx14}{1.00000}

\calculatefx{oracle}{1}
\draw[thick, draw=Fuchsia](\rotate+360/\n:\csname oracle\endcsname)--(\rotate+360/\n*2:\csname oracle\endcsname)--(\rotate+360/\n*3:\csname oracle\endcsname)--(\rotate+360/\n*4:\csname oracle\endcsname)--(\rotate+360/\n*5:\csname oracle\endcsname)--(\rotate+360/\n*6:\csname oracle\endcsname)--(\rotate+360/\n*7:\csname oracle\endcsname)--cycle;

\draw[fill=Fuchsia,thick,opacity=0.05](\rotate+360/\n:\csname oracle\endcsname)--(\rotate+360/\n*2:\csname oracle\endcsname)--(\rotate+360/\n*3:\csname oracle\endcsname)--(\rotate+360/\n*4:\csname oracle\endcsname)--(\rotate+360/\n*5:\csname oracle\endcsname)--(\rotate+360/\n*6:\csname oracle\endcsname)--(\rotate+360/\n*7:\csname oracle\endcsname)--cycle;
\draw[fill=RoyalBlue,thick,opacity=0.05](\rotate+360/\n:\csname fx8\endcsname)--(\rotate+360/\n*2:\csname fx9\endcsname)--(\rotate+360/\n*3:\csname fx10\endcsname)--(\rotate+360/\n*4:\csname fx11\endcsname)--(\rotate+360/\n*5:\csname fx12\endcsname)--(\rotate+360/\n*6:\csname fx13\endcsname)--(\rotate+360/\n*7:\csname fx14\endcsname)--cycle;
\draw[fill=Orange,thick,opacity=0.15](\rotate+360/\n:\csname fx1\endcsname)--(\rotate+360/\n*2:\csname fx2\endcsname)--(\rotate+360/\n*3:\csname fx3\endcsname)--(\rotate+360/\n*4:\csname fx4\endcsname)--(\rotate+360/\n*5:\csname fx5\endcsname)--(\rotate+360/\n*6:\csname fx6\endcsname)--(\rotate+360/\n*7:\csname fx7\endcsname)--cycle;

\draw[thick,draw=RoyalBlue](\rotate+360/\n:\csname fx8\endcsname)--(\rotate+360/\n*2:\csname fx9\endcsname)--(\rotate+360/\n*3:\csname fx10\endcsname)--(\rotate+360/\n*4:\csname fx11\endcsname)--(\rotate+360/\n*5:\csname fx12\endcsname)--(\rotate+360/\n*6:\csname fx13\endcsname)--(\rotate+360/\n*7:\csname fx14\endcsname)--cycle;
\draw[thick,draw=Orange](\rotate+360/\n:\csname fx1\endcsname)--(\rotate+360/\n*2:\csname fx2\endcsname)--(\rotate+360/\n*3:\csname fx3\endcsname)--(\rotate+360/\n*4:\csname fx4\endcsname)--(\rotate+360/\n*5:\csname fx5\endcsname)--(\rotate+360/\n*6:\csname fx6\endcsname)--(\rotate+360/\n*7:\csname fx7\endcsname)--cycle;

\draw(\rotate+360/\n:\N*\Size+1.5)node[text width=1cm,align=center]{\footnotesize PACS\\[-4pt]ResNet18}; 
\draw(\rotate+2*360/\n:\N*\Size+2)node[text width=1cm,align=center]{\footnotesize PACS\\[-4pt]ViT-S};  
\draw(\rotate+3*360/\n:\N*\Size+2)node[text width=1cm,align=center]{\footnotesize MiniDN\\[-4pt]AlexNet};  
\draw(\rotate+4*360/\n:\N*\Size+1.6)node[text width=1cm,align=center]{\footnotesize MiniDN\\[-4pt]ResNet18};
\draw(\rotate+5*360/\n:\N*\Size+1.6)node[text width=1cm,align=center]{\footnotesize Digits\\[-4pt]LeNet};
\draw(\rotate+6*360/\n:\N*\Size+2.1)node[text width=1cm,align=center]{\footnotesize Cam17\\[-4pt]ResNet50};
\draw(\rotate+7*360/\n:\N*\Size+2.2)node[text width=1cm,align=center]{\footnotesize NICO++\\[-4pt]ResNet18};

\draw(\rotate+360/\n:\csname oracle\endcsname + 0.5)node[text width=1cm,align=center]{\tiny 66.2}; 
\draw(\rotate+2*360/\n:\csname oracle\endcsname + 0.5)node[text width=1cm,align=center]{\tiny 75.7};  
\draw(\rotate+3*360/\n:\csname oracle\endcsname + 0.5)node[text width=1cm,align=center]{\tiny 49.1};  
\draw(\rotate+4*360/\n:\csname oracle\endcsname + 0.5)node[text width=1cm,align=center]{\tiny 57.9};
\draw(\rotate+5*360/\n:\csname oracle\endcsname + 0.5)node[text width=1cm,align=center]{\tiny 83.8};
\draw(\rotate+6*360/\n:\csname oracle\endcsname + 0.5)node[text width=1cm,align=center]{\tiny 94.5};
\draw(\rotate+7*360/\n:\csname oracle\endcsname + 0.5)node[text width=1cm,align=center]{\tiny 29.1};

\draw(\rotate+360/\n:\csname fx8\endcsname - 0.5)node[text width=1cm,align=center]{\tiny 65.9}; 
\draw(\rotate+2*360/\n:\csname fx9\endcsname - 0.5)node[text width=1cm,align=center]{\tiny 74.5};  
\draw(\rotate+3*360/\n:\csname fx10\endcsname - 0.5)node[text width=1cm,align=center]{\tiny 48.9};  
\draw(\rotate+4*360/\n:\csname fx11\endcsname - 0.5)node[text width=1cm,align=center]{\tiny 57.4};
\draw(\rotate+5*360/\n:\csname fx12\endcsname - 0.5)node[text width=1cm,align=center]{\tiny 82.6};
\draw(\rotate+6*360/\n:\csname fx13\endcsname - 0.5)node[text width=1cm,align=center]{\tiny 93.6};
\draw(\rotate+7*360/\n:\csname fx14\endcsname - 0.5)node[text width=1cm,align=center]{\tiny 29.1};

\draw(\rotate+360/\n:\csname fx1\endcsname - 0.3)node[text width=1cm,align=center]{\tiny 51.8}; 
\draw(\rotate+2*360/\n:\csname fx2\endcsname - 0.6)node[text width=1cm,align=center]{\tiny 68.7};  
\draw(\rotate+3*360/\n:\csname fx3\endcsname - 0.4)node[text width=1cm,align=center]{\tiny 39.8};  
\draw(\rotate+4*360/\n:\csname fx4\endcsname - 0.6)node[text width=1cm,align=center]{\tiny 53.0};
\draw(\rotate+5*360/\n:\csname fx5\endcsname - 0.3)node[text width=1cm,align=center]{\tiny 72.5};
\draw(\rotate+6*360/\n:\csname fx6\endcsname - 0.4)node[text width=1cm,align=center]{\tiny 78.7};
\draw(\rotate+7*360/\n:\csname fx7\endcsname - 0.6)node[text width=1cm,align=center]{\tiny 26.9};

\draw [thick, RoyalBlue] (-9,-8.5) -- (-8.3,-8.5); 
\node at (-6,-8.5) {Proposed $V_A$ };
\draw [thick, Orange] (-2,-8.5) -- (-1.3,-8.5); 
\node at (0.8,-8.5) {Standard $V_S$ };
\draw [thick, Fuchsia] (4.8,-8.5) -- (5.5,-8.5); 
\node at (7.5,-8.5) {Oracle $V_O$};
\end{tikzpicture}
\vspace{-5pt}
\caption{
\textbf{Test accuracy comparison of different validation methods across seven dataset-backbone configurations.}
Each axis ranges from $70\%$ to $100\%$ of the oracle validation $V_O$ that assumes access to the test set. The proposed augmented validation $V_A$ achieves over $\boldsymbol{98\%}$ of the oracle performance on average, representing a $\boldsymbol{15.4\%}$ relevant improvement over the standard validation $V_S$. For each dataset, the model is chosen from a pool, with $4,500$ trained models across all pools.
\label{fig:radar}
}
\vspace{-15pt}
\end{figure}

Visual recognition models are primarily trained using data from one or multiple \emph{source} domains, typically the richest in labeled data or the only available domains during training. The ability of a visual recognition model to generalize or adapt to novel \emph{target} domains with no or limited labeled data is a desirable property. The literature explores such generalization and adaptation under various setups~\cite{wkt+16, c17}. Supervised domain adaptation refers to cases where labeled examples from the target domain are available, while unsupervised domain adaptation deals with unlabeled examples. The task is called domain generalization when there is a complete lack of target domain examples. Depending on whether there is a single or multiple source domains, the task is categorized as single-source domain generalization (SSDG) or multi-source domain generalization (MSDG), respectively.

The lack of labeled target domain examples is a challenging factor for performing both model validation, \ie, estimating a model's accuracy to tune its hyper-parameters, and method selection, \ie, evaluating which method is the best. This often leads to oracle-based validation, where access to the test set is incorrectly assumed. This aspect is investigated in the work of Musgrave \etal~\cite{mbl21} for unsupervised domain adaptation and in Gulrajani \etal~\cite{gl20} for MSDG, where proper validation protocols are suggested to avoid malpractice. 
In the context of MSDG, vanilla training on the source domains is top-performing~\cite{gl20} when methods are tested on datasets they were not designed for. Therefore, effective method selection processes are of paramount importance in practice.
\looseness=-1

In the context of SSDG, developing a visual recognition model robust to unseen distribution shifts during test time is the most challenging of the tasks above. 
Due to the absence of a proper validation set and protocol, practitioners need to rely on educated guesses to enhance the generalizability of their models. Without an effective validation protocol, it becomes impossible to determine the effectiveness of any model enhancement. 
To our knowledge, no prior work studies validation protocols in SSDG.

\begin{figure}
\centering
\resizebox{1\columnwidth}{!}{%
\setlength{\fboxsep}{0pt}%
\setlength{\fboxrule}{0.5pt}%
\scriptsize
\begin{tabular}{c@{}c@{}c@{}c@{}}
\fcolorbox{black}{black}{\includegraphics[height=1.4cm]{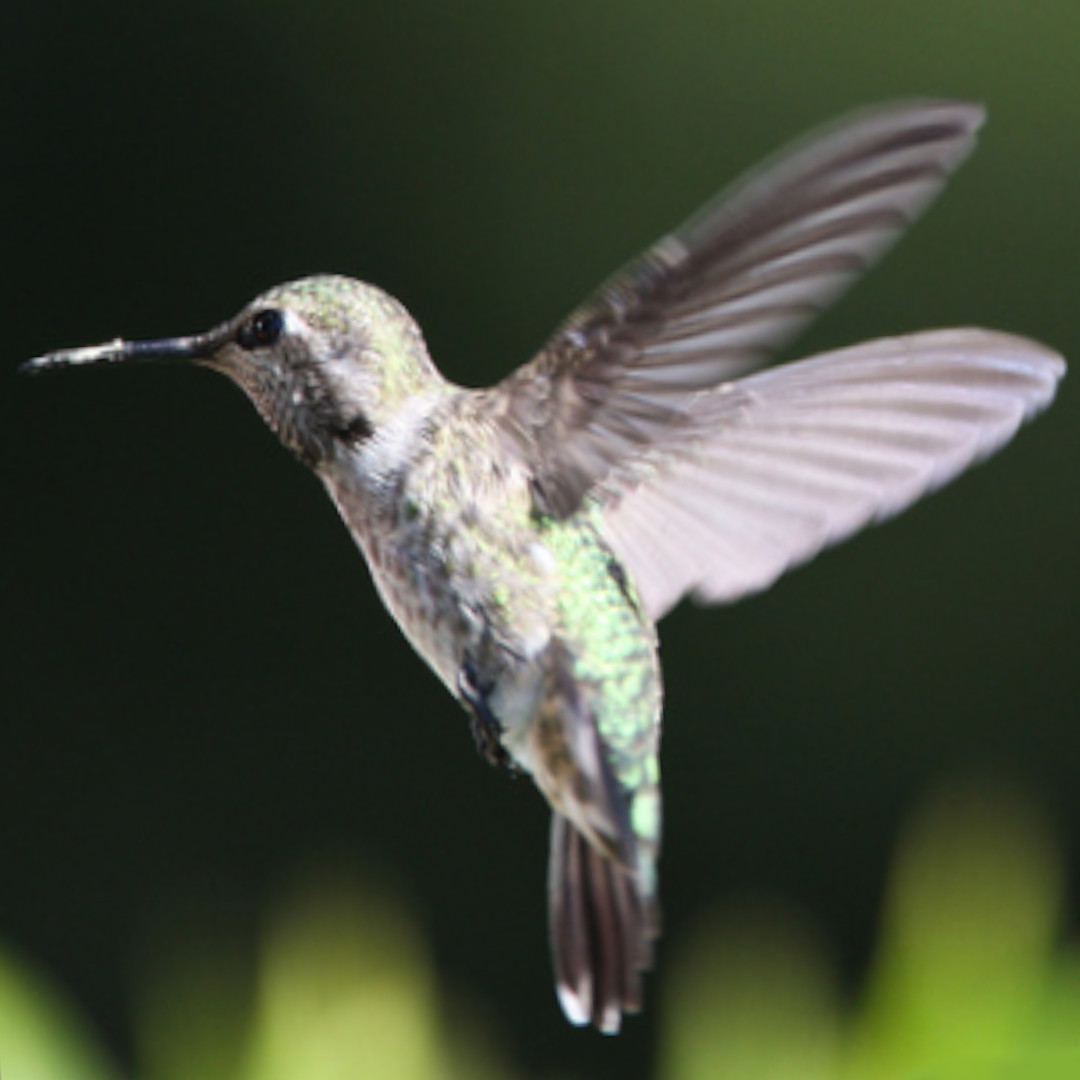}}\hspace{0.1cm} &
\fcolorbox{black}{black}{\includegraphics[height=1.4cm]{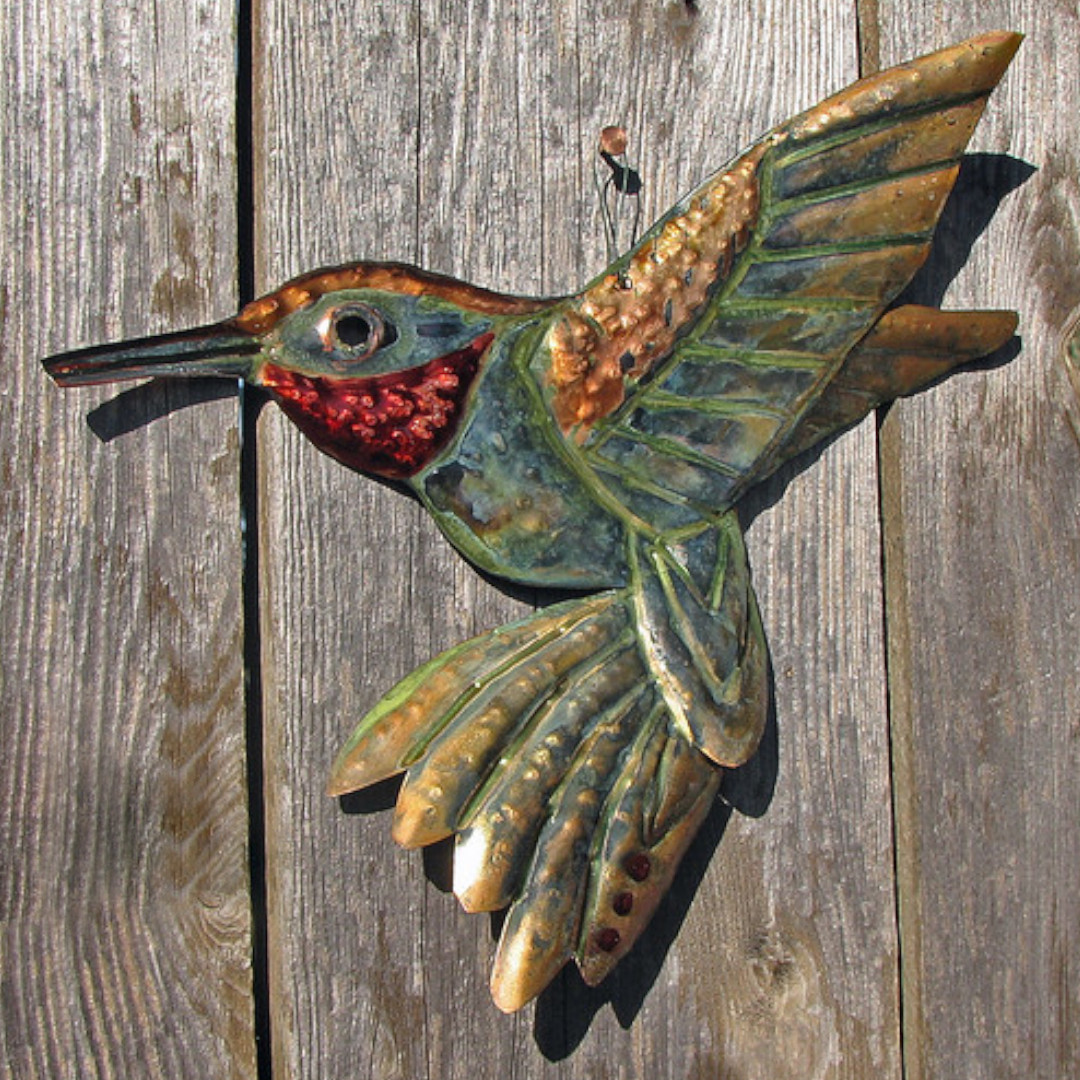}}\hspace{0.1cm} &
\fcolorbox{black}{black}{\includegraphics[height=1.4cm]{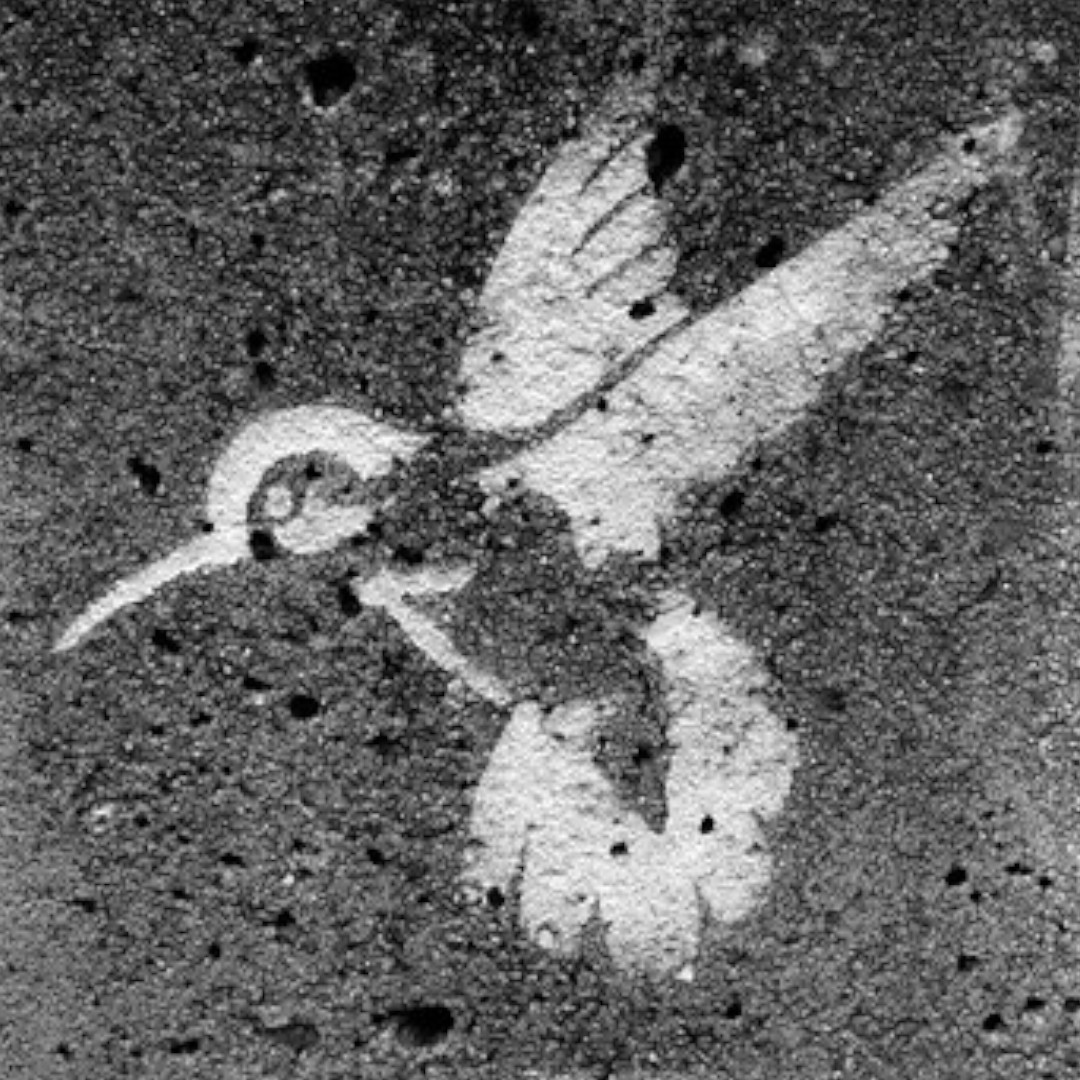}}\hspace{0.1cm} &
\fcolorbox{black}{black}{\includegraphics[height=1.4cm]{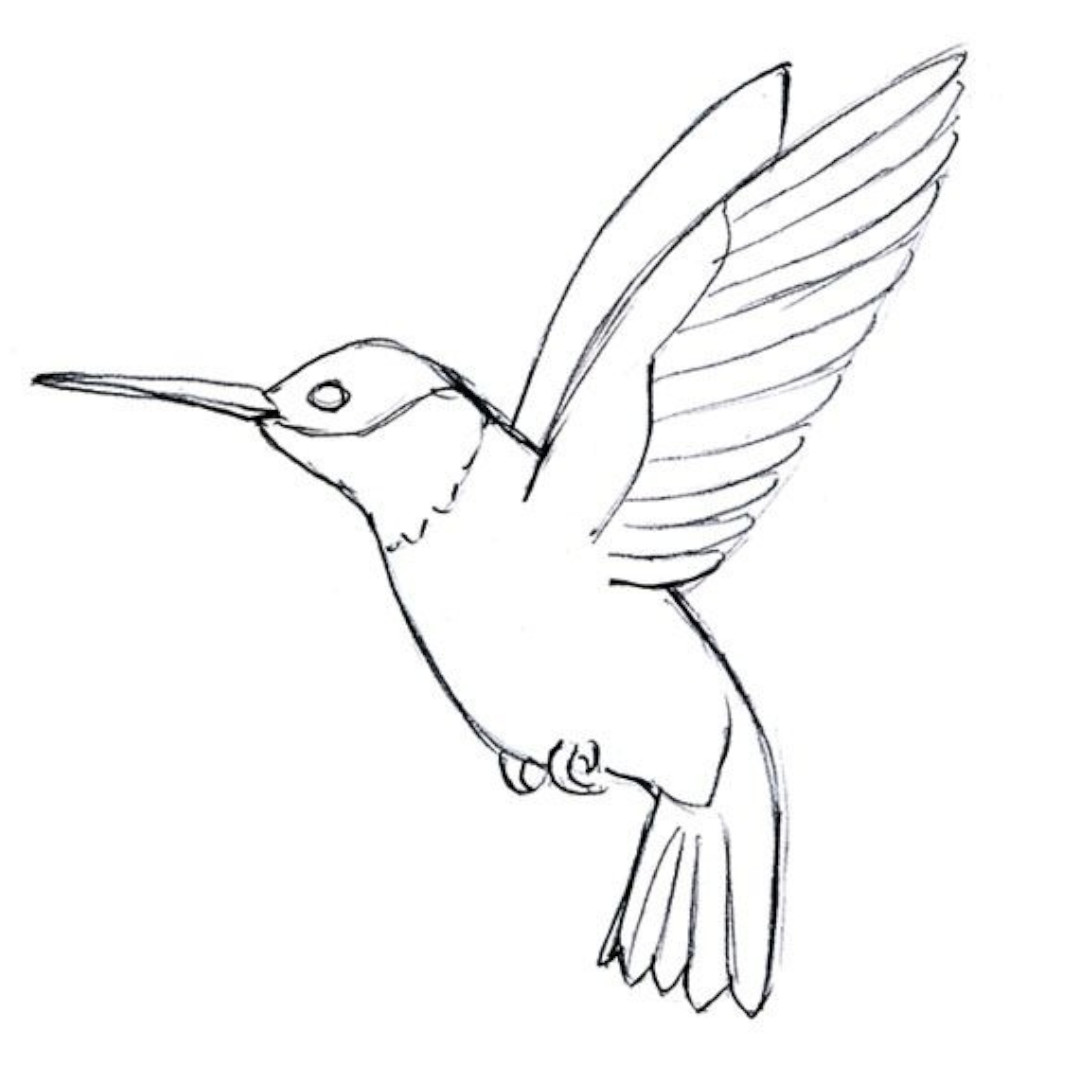}}\hspace{0.1cm}
\\[-0.02cm]
photo &  sculpture  &  graffiti & sketch \\

\end{tabular}

}

\vspace{-5pt}
    \caption{``Hummingbird'' in four different domains. The domain generalization task tacitly assumes domains that share informative, human-understandable features, such as texture, shape, or semantics. Therefore, the unseen domains are expected to be human-recognizable. Images from ImageNet \cite{dsl+09} and ImageNet-R \cite{imgnetr}}
    \vspace{-15pt}
    \label{fig:domainexamples}
\end{figure}

In this work, we follow a fundamental direction in the generalization task, \ie, data augmentations, to synthesize new distributions. The effect of data augmentations is studied extensively in the context of domain generalization, both in combination with adversarial learning \cite{ADA, Wlq+21} and standalone \cite{RandConv, ProRandConv}. However, in prior work, augmentations were exclusively used to extend the training set in size and variability. Instead, we apply them to source domain images to obtain a validation set with increased variability and estimate the method's performance on unseen distributions.
We argue that exploiting an exhaustive list of existing augmentations synthesizes instances that span various human-recognizable domains. These augmentations preserve human-recognisable features by their design, while the attempt for completeness ensures the coverage of several unseen potential domains (Figure~\ref{fig:domainexamples}). Since these augmentations are also valuable in the training phase, we propose a $k$-fold cross-validation scheme performed across augmentation types to get the best of both worlds. This way, the training set is augmented with challenging examples while, at the same time, the validation provides an unbiased estimate of performance on unseen distributions. Using the proposed validation for model selection and hyper-parameter tuning, we achieve better performance than the standard practice on various datasets (see Figure~\ref{fig:radar}).
\looseness=-1

Besides the novel validation method for SSDG, we propose a family of classification methods parametrized by several train and test-time hyper-parameters. The values of these parameters are selected by the proposed validation method. We focus on enforcing shape bias~\cite{grm+19}, whose effectiveness is demonstrated in prior work~\cite{nrk21, hsa+22, nk22}. We accomplish this by using a specialized image transformation technique, employing enhanced edge maps that eliminate textures while retaining crucial shape information. The transformation is performed both during training and testing, with and without randomization, respectively. Despite its simplicity, the proposed method sets a new state-of-the-art on multiple benchmarks, highlighting the value of pronounced shape information and exhaustive augmentations, as well as the effectiveness of the proposed validation (see Figure~\ref{fig:all_datasets}). 

\looseness=-1
\section{Related work}
\label{sec:related}
We review the prior work on MSDG and SSDG, a less explored task, and discuss shape bias methods in domain generalization and prior work on augmentation strategies.

\paragraph{Multi-source domain generalization.} Domain-invariant feature learning is the most popular family of approaches, originating from the results of Ben-David \etal~\cite{bbc+07}: The upper bound of the target domain error is a function of the discriminability between the source and target domain in the feature space. Many follow-up methods exist in the MSDG literature, such as kernel-based approaches~\cite{mbs13} or multi-task autoencoders~\cite{gbz+15}. Ganin \etal ~\cite{gua+16} and Li \etal ~\cite{ltg+18} perform adversarial learning to match the domain distributions. Kim \etal ~\cite{SelfReg} propose bringing the same-class representations closer regardless of domain. Li \etal \cite{lzy+19} encourage domain invariance by mixing domain-specific network components with domain-agnostic ones.

The use of data augmentations during training is another common approach. Zhou \etal ~\cite{zyh+20} synthesize new examples from pseudo-novel domains conditioned on existing examples while enforcing semantic consistency. Stylization using images from different domains as styles is a simple but effective approach in the work of Somavarapu \etal~\cite{smk20}, while Mancini \etal ~\cite{mar+20} use mixup~\cite{zcd+18} to combine different source domains. 
Carlucci \etal ~\cite{JiGen} additionally optimize the classification loss to train a model that solves jigsaw puzzles in a self-supervised manner. Mansilla \etal~\cite{mem21} identify and control domain-specific conflicting gradients.

Regarding generalization in real-world domain shifts, two popular benchmarks exist for MSDG: iWildCam~\cite{wilds} and NICO++~\cite{nico++}. In iWildCam, a tiny fraction of all classes are seen in each domain, making this benchmark unsuitable for SSDG. NICO++, on the other hand, consists of real photographs taken in different conditions and is suitable for SSDG. We repurpose it for our task to demonstrate that both the proposed method and validation protocol improve under real-world domain shifts.

\paragraph{Single-source domain generalization.}
Most existing SSDG methods rely on data augmentation or data generation. Volpi \etal ~\cite{ADA} generate hard examples from an imaginary target domain, while Yue \etal ~\cite{yzz+19} use style transfer to produce images of novel styles. Qiao \etal ~\cite{qzp20} use adversarial domain augmentation and an auxiliary Wasserstein autoencoder to force semantic consistency between the augmented and original images in the latent space. Xu \etal~\cite{RandConv} propose random convolutions as an augmentation technique to diversify the training data. Wang \etal ~\cite{Wlq+21} propose a style-complement module to transform training examples in a way that is complementary to the source domain. 
A Fourier-based augmentation mixing the amplitude of two images is proposed by Xu \etal~\cite{xzz21}, who assume that the Fourier phase is not easily affected by domain shifts. Wan \etal~\cite{wsz22} target domain invariance through a decomposition and composition technique that builds on the bag-of-words model. Lee \etal~\cite{lkk22} use a distillation approach by creating an ensemble prediction from images of the same class and penalizing the mismatched outputs with the ensemble. Chen \etal~\cite{CADA} propose a center-aware adversarial augmentation that enriches the training samples by pushing them away from the class centers using an angular center loss. Chen \etal~\cite{ITTA} use a learnable consistency loss for test-time adaptation, and they introduce additional adaptive parameters during the test phase. Chen \etal~\cite{MCL} propose a new learning paradigm for training with domain shifts by employing meta-causal learning to simulate a domain shift, analyze its causes, and reduce it. Choi \etal~\cite{ProRandConv} enhance the idea of random convolution by recursively stacking ones with small kernel sizes, deformable offsets, and affine transformations.

\paragraph{Shape bias.} Geirhos \etal ~\cite{grm+19} show that, in contrast to human subjects, CNNs trained on ImageNet are biased to focus on textures and mitigate that by training on  a stylized version of ImageNet. SagNet~\cite{SagNet} disentangles style encodings from class categories to prevent style bias and to focus more on object shapes. Edge detection as a bridge between domains has a prominent role, such as in the work of Harary \etal~\cite{hsa+22}, who target few-shot learning and rely on domain labels. Another example is the work of Nazari and Kovashka~\cite{nk22}, where edge detection forms a fixed augmentation used both for training and testing. Our work has similarities but also differences. Namely, we use improved shape representations, while our method uses a single backbone instead of one backbone for images and one for edge maps. The superiority of this choice is experimentally validated. Narayanan \etal ~\cite{nrk21} argue that the shock graph of the contour map of an image is a complete representation of its shape content. As a drawback, the high cost of their approach does not allow pre-training on ImageNet, and any corresponding performance gain is lost. 

\paragraph{Augmentation strategies.} Various data augmentation techniques exist to enhance model performance. Common practices include flipping and cropping~\cite{hzr+16}, occlusion (Cutout)~\cite{cutout}, segment replacement (CutMix)~\cite{cutmix}, or element-wise convex combination of images (Mixup)~\cite{zcd+18}. Learned approaches, like AutoAugment\cite{AutoAugment}, tune an augmentation set to optimize the performance of downstream tasks. Alternatively, Patch Gaussian \cite{patchgaussian} applies Gaussian noise to random image portions as an augmentation. AugMix~\cite{AugMixpaper} combines randomly generated augmentations while ensuring consistency through a Jensen-Shannon loss. In this work, we randomly sample augmentations from a single library to avoid introducing bias by tailoring the selection to specific datasets. Nevertheless, the above strategies can be orthogonally used with the proposed augmented validation protocol.
\section{Training, validation and testing in SSDG}
\label{sec:method}
\begin{figure*}[t]
\begin{center}
\includegraphics[width=0.95\textwidth]{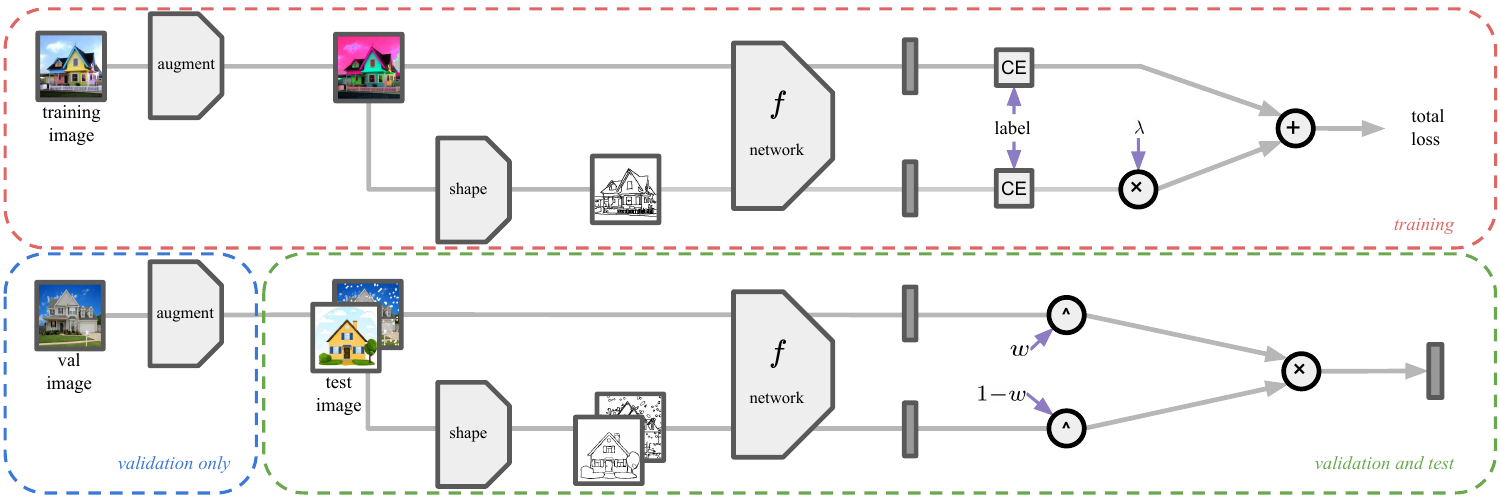}
\vspace{-5pt}
\caption{
\textbf{Overview of the training, validation, and testing pipeline.} Training images are augmented with \emph{basic} and a sub-set of \emph{extra} augmentations. The shape information, encoded as binary thin edges of the augmented image, generates an additional training example. The contribution of the two losses, image and shape-based, is weighed by a parameter $\lambda$. 
In validation, extra augmentations that were not included in the training are used to synthesize unseen distributions. The shape information is optionally exploited in testing and in the validation phase. The final prediction is obtained by ensembling the image and the shape-based predictions, weighted by a parameter $w$.
\label{fig:teaser}
\vspace{-25pt}
}
\end{center}
\end{figure*}

SSDG training is performed on a source domain, and the model is tested on an unseen target domain. The same categories are to be recognized in both source and target domains, \ie, the label spaces are identical. The input space of the source and target domain are RGB images, but their distributions differ. 
The goal is to perform training so the model can generalize to the target domain. A validation set is necessary but can only be constructed from images of the source domain. Nevertheless, using the raw images is unlikely to reflect any generalization ability in the validation.

To evaluate the generalization capabilities of a model, we introduce the concept of an augmented validation set. In contrast to the well-established practice of using a small set of domain-appropriate augmentations to training images, a wide range of augmentations are drawn from an exhaustive list of existing techniques. This validation set is used in conjunction with the proposed SSDG method, whose key component is the direct injection of shape information in the training procedure. The augmentations used for the validation set are also optionally used in the training. In that case, a two-fold cross-validation process over the augmentation groups is proposed to ensure the validation is performed on a previously unseen distribution. 

During the testing phase, two possible approaches are explored. The first performs classification on the input image alone. The second leverages both the input image and its corresponding shape information -- the weighted average of the predictions from both inputs is used (see Figure \ref{fig:teaser}).
\looseness=-1

\subsection{Proposed validation set}

\paragraph{Motivation:} 
Proper validation is crucial for effective model selection and hyperparameter tuning in any machine learning task, including SSDG. In the context of SSDG, the standard practice is to employ a validation set from raw images of the source domain. However, this validation set alone cannot accurately reflect the model's generalization performance across other domains. As a result, methods that perform well in the source domain are favored by such a process.
We introduce a synthetic distribution shift by manipulating the validation set of the source domain to evaluate performance under a distribution shift. This allows us to emulate the challenges posed by domain shifts and obtain a more realistic assessment of the model's generalization capabilities.
To capture the generalization performance across multiple domains, we heavily rely on existing image augmentations. Specifically, we incorporate a wide range of diverse augmentations into our approach, aiming to cover as many variations as possible. 

\paragraph{Variants:}
The following constructions of the validation set are considered and compared.
\vspace{-4pt}
\begin{itemize}[leftmargin=10pt]
\setlength\itemsep{-4pt}
\item[$\bullet$] \paragraph{Oracle - $V_O$:} The \emph{Oracle validation set} is equivalent to the test set. In the case of multiple test sets, their union is considered. 
In this work, oracle validation is {\bf never} used to compare performance with prior approaches. It is only used as an upper bound on method performance and to measure how close the results achieved with the proposed validation set/method are to the best possible.
\item[$\bullet$] \paragraph{Standard - $V_S$:} This validation set consists of raw images from the source domain. In this case, there is no distribution shift between the training and validation sets. It is also referred to as the \emph{standard validation set} and is equivalent to validation on the training distribution.
\item[$\bullet$] \paragraph{Augmented - $V_A$:} 
A total of 76 different augmentations organized into 10 groups according to their types are used to alter the images. Each image in the standard validation set $V_S$ is augmented 10 times by one random augmentation per group, resulting in a validation set 10 times larger than $V_S$. This \emph{augmented validation set} is created once. The same set is used in all experiments for $V_A$. The ImgAug\footnote{\url{https://github.com/aleju/imgaug}} library is utilized with all its implemented augmentations and category groups. We refer to these as \emph{extra augmentations} to differentiate them from the \emph{basic augmentations} (random crop, scale, and flip).

\end{itemize}

\subsection{Proposed recognition method}

\paragraph{Motivation:} 
The proposed family of methods builds upon previous findings that highlight the texture bias in CNNs~\cite{grm+19}. Since CNNs primarily capture texture cues during the training, their performance tends to suffer when confronted with domains lacking these texture cues. To address this limitation, we introduce an explicit enhancement of the shape bias by incorporating shape information extracted from the images in the form of edges.
Shape is a fundamental characteristic across multiple domains, making it valuable for bridging the domain gap. By using edge detection, we enable the mapping of objects to a common domain. The shape information is used both during training and testing to reinforce this bridging effect. The parameters mixing the image and shape information are tuned according to the augmented validation. 

We additionally investigate the effect of using the extra augmentations during training. Overlap between training and validation distributions may result in overestimating the expected performance due to validation on a seen distribution. To avoid this, the training and validation augmentation groups are kept disjointed via a two-fold validation process.

\paragraph{Network architecture:} A standard backbone taking an RGB image as input followed by a linear classifier providing a class probability at the output is employed. In our experiments, AlexNet~\cite{ksh12}, ResNet~\cite{hzr+16}, and ViT~\cite{vit} are used.

\paragraph{Shape extraction:} To obtain an explicit shape representation, a modified version of the binary thin edges (BTE)~\cite{etc22} is used. BTE maps the input image to a binary edge map. Edges are extracted by the Sobel operator instead of the learned edge detectors used in~\cite{etc22}. Detected edges are further processed by thinning, hysteresis with an adaptive threshold, and removing small connected components. During training, shape variation is achieved with randomized geometric augmentation and randomization of the thresholding process.
Binarization provides a form of cleaning shape-irrelevant information, typically corresponding to background.

\paragraph{Training augmentations:} \emph{Basic augmentations} (random crop, scale, and flip) are used in all our experiments. When the \emph{extra augmentations} are used in training, each input image is transformed with at most one randomly sampled extra augmentation performed before the basic augmentations.

\paragraph{Training variants:} A family of methods is proposed and experimentally evaluated. The network is always trained with cross-entropy loss, while  the network input during the training phase varies.
\vspace{-4pt}
\begin{itemize}[leftmargin=10pt]
\setlength\itemsep{-4pt}
\item[$\bullet$] \paragraph{$I$:} Training baseline - training is performed on images of the source domain.
\item[$\bullet$] \paragraph{$\hat{I}$:} Training is performed on images of the source domain with extra augmentations.
\item[$\bullet$] \paragraph{$S$:} Training is performed with shape information only. Shape is captured in the form of BTEs obtained from images of the source domain.
\item[$\bullet$] \paragraph{$IS$:} Training is performed with both images of the source domain and their BTEs. The loss on the image $\ell_I$ and the loss on the shape information $\ell_S$ are combined by $\ell_I + \lambda \ell_S$. Unless otherwise stated, the two losses are balanced by $\lambda=1.0$.
\item[$\bullet$] \paragraph{$\hat{I}S$:} Same as the above but with extra augmentations applied to the input images. 
\item[$\bullet$] \paragraph{$IS_{\times 2}$:} Same as $IS$ but with two separate backbones for images and BTEs.
\end{itemize}

\paragraph{Testing variants:}  The following variants of the network input during test time are considered.
\vspace{-4pt}
\begin{itemize}[leftmargin=10pt]
\setlength\itemsep{-4pt}
\item[$\bullet$] \paragraph{$I$:} Testing baseline -  testing is performed on the input images of the target domain.
\item[$\bullet$] \paragraph{$S$:} Testing is performed on BTEs of the target domain.
\item[$\bullet$] \paragraph{$IS$:} Testing is performed on both images and BTEs of the target domain. Let's denote by $p_I$ and $p_S$ the probability of a particular class based on the input image and BTE, respectively. The combined response is given by their geometric mean $p_I^w p_S^{(1-w)}$, with $w\in [0,1]$. This is a generalized approach that reduces to testing solely on $I$ or $S$ for $w=1$ or $w=0$, respectively. When reporting the value of $w$, we denote this variant by $I^w S$ for brevity.
\item[$\bullet$] \paragraph{$IS_{\times 2}$:} Both backbones, each with the corresponding input, are used for testing, which aims to form a more costly approach for the experimental comparisons.
\end{itemize}

\paragraph{Training-testing combinations:} The training-testing combinations are denoted by $A\righttarrow B$ for training with variant $A$ and testing with variant $B$. The \emph{baseline} method that trains on source domain images (training baseline) and tests on target domain images (testing baseline) is denoted as $I\righttarrow I$. In principle, all combinations are possible since there is no testing constraint, only on inputs during training. Nevertheless, we mostly focus on the following: $\hat{I}\righttarrow I$, $S\righttarrow S$, $IS \righttarrow IS$, $\hat{I}S \righttarrow IS$, and $IS_{\times 2} \righttarrow IS_{\times 2}$. An overview of our training, validation, and testing process is shown in Figure~\ref{fig:teaser}.

\paragraph{Two-fold cross-validation over augmentation groups:} Training with the extra augmentations, \ie, $\hat{I}$ variants, which are used for the validation set, also invalidates the concept of validating on unseen distributions. Therefore, we propose a two-fold cross-validation (TCV) protocol. That is, to train using half of the augmentation groups while keeping the rest to create the validation set. This process is repeated twice, once for each half used in training, and the validation performance is the average of the two runs for a particular variant or value of a hyperparameter. During the final stage, the network is trained with the chosen hyper-parameters using all available augmentations.
\section{Experiments}
\label{sec:exp}

\subsection{Experimental setup and implementation}

The proposed validation and classification methods are evaluated on five datasets, namely Digits~\cite{lbb+98, nwc+11, DANN, dgg+88}, PACS~\cite{lys+17}, Mini-DomainNet~\cite{pbx19}, NICO++~\cite{nico++}, and Camelyon17~\cite{camelyon}. The classifier performance is measured by \emph{classification accuracy}.
Digits is composed of five different datasets, namely MNIST~\cite{lbb+98}, SVHN~\cite{nwc+11}, MNIST- M~\cite{DANN}, SYN~\cite{DANN}, and USPS~\cite{dgg+88}, each one corresponding to a different domain. PACS is a domain generalization dataset with four domains: photo, art paintings, cartoon, and sketch. Mini-DomainNet is a subset of the domain generalization dataset DomainNet~\cite{pbx19}. It consists of four domains: clipart, painting, real, and sketch. NICO++ consists of natural images, with domains defined as the context: autumn, dim light, grass, outdoor, rock, and water. Camelyon17 is a medical tumor detection dataset, and the domains are defined by the five different hospitals that provided the data. We follow the setup of \cite{ABA} and train on hospitals $1$, $2$, and $3$ without using the domain labels, while we test on hospitals $4$ and $5$.  
The source domain is set to be MNIST, \emph{photos}, and \emph{real} for Digits, PACS, and Mini-DomainNet, respectively. We are the first to work on SSDG with NICO++, where we treat each domain as a source domain and the five others as target domains.

We use AlexNet~\cite{ksh12}, ResNet-18, and ResNet-50~\cite{hzr+16} pre-trained on ImageNet-1k~\cite{dsl+09}, and ViT-Small~\cite{vit} pre-trained first on ImageNet-21k~\cite{imagenet21k} and then on ImageNet-1k. The pre-trained networks justify using only \emph{photos} and \emph{real} as source domains for PACS and Mini-DomainNet; all other options violate the SSDG protocol since two domains are seen during training. On Digits we use a simple LeNet (conv-relu-pool-conv-relu-pool) architecture~\cite{JiGen, ADA}. On NICO++, following the dataset guidelines, we start with a randomly initialized ResNet-18. Implementation details are included in the Appendix.

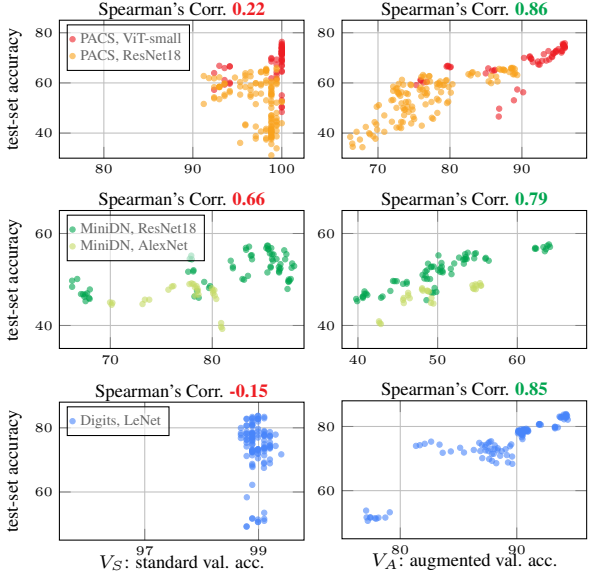
\begin{figure}[t]
\centering
\renewcommand{\arraystretch}{0.7}%
\begin{tabular}{c@{}c@{}}
\pgfplotsset{every tick label/.append style={font=\tiny}}
\begin{tikzpicture}[define rgb/.code={\definecolor{mycolor}{RGB}{#1}}, rgb color/.style={define rgb={#1},mycolor}]
\scriptsize
\begin{axis}[
    title style={yshift=-5pt,},
    title={Spearman's Corr. \red{\textbf{0.22}}},
    width=0.58\linewidth,
    height=0.41\linewidth,
    xmin=75.0,
    xmax=102.5,
    ymin=30.0,
    ymax=85.0,
    grid=both,
    xtick={80, 90, 100},
    ytick={40,60, 80},
    ylabel = {test-set accuracy},
    xlabel style={yshift=5pt},
        title style={yshift=-3pt},
    legend pos=north west,
    legend style={cells={anchor=west}, font =\scriptsize, fill opacity=0.6, row sep=-2.5pt,inner sep=1pt},
    ]
\addplot[only marks, mark=*, opacity=0.6,mark size=1.0, color=Red, error bars/.cd, 
    y fixed,
    y dir=both, 
    y explicit] table[x=Source_val,y=Source_PACS_ViT
]{./data/main/total_scatter.csv};
\addlegendentry{\scalebox{.8}{PACS, ViT-small}};
\addplot[only marks, mark=*, opacity=0.6 ,mark size=1.0, color=YellowOrange, error bars/.cd, 
    y fixed,
    y dir=both, 
    y explicit] table[x=Source_val,y=Source_PACS_ResNet
]{./data/main/total_scatter.csv};
\addlegendentry{\scalebox{.8}{PACS, ResNet18}};
\end{axis}
\end{tikzpicture}

&

\pgfplotsset{every tick label/.append style={font=\tiny}}
\begin{tikzpicture}[define rgb/.code={\definecolor{mycolor}{RGB}{#1}}, rgb color/.style={define rgb={#1},mycolor}]
\scriptsize
\begin{axis}[
    title style={yshift=-5pt,},
    title={Spearman's Corr. \green{\textbf{0.86}}},
    width=0.58\linewidth,
    height=0.41\linewidth,
    xmin=65.0,
    xmax=99,
    ymin=30.0,
    ymax=85.0,
    ytick={40,60,80},
    xlabel style={yshift=5pt},
    grid=both,
    xtick={70, 80, 90},
    title style={yshift=-3pt},
    legend pos=north west,
    legend style={cells={anchor=west}, font=\scriptsize, fill opacity=0.8, row sep=-2.5pt},
    ]
\addplot[only marks, mark=*, opacity=0.6,mark size=1.0, color=Red, error bars/.cd, 
    y fixed,
    y dir=both, 
    y explicit] table[x=ImgAug_val,y=ImgAug_PACS_ViT
]{./data/main/total_scatter.csv};
\addplot[only marks, mark=*, opacity=0.6 ,mark size=1.0, color=YellowOrange, error bars/.cd, 
    y fixed,
    y dir=both, 
    y explicit] table[x=ImgAug_val,y=ImgAug_PACS_ResNet
]{./data/main/total_scatter.csv};
\end{axis}
\end{tikzpicture}

\\ 

\pgfplotsset{every tick label/.append style={font=\tiny}}
\begin{tikzpicture}
\scriptsize
\begin{axis}[
    title style={yshift=-5pt,},
    title={Spearman's Corr. \red{\textbf{0.66}}},
    width=0.58\linewidth,
    height=0.41\linewidth,
    xmin=65.0,
    xmax=89,
    ymin=35.0,
    ymax=65.0,
    grid=both,
    xtick={70, 80},
    ytick={40,60},
    xlabel style={yshift=5pt},
    ylabel = {test-set accuracy},
    title style={yshift=-3pt},
    legend pos=north west,
    legend style={cells={anchor=west}, font =\scriptsize, fill opacity=0.6, row sep=-2.5pt,inner sep=1pt},
    ]
\addplot[only marks, mark=*, opacity=0.6 ,mark size=1.0, color=Green, error bars/.cd, 
    y fixed,
    y dir=both, 
    y explicit] table[x=Source_val,y=Source_MiniDN_ResNet
]{./data/main/total_scatter.csv};
\addlegendentry{\scalebox{.8}{MiniDN, ResNet18}};
\addplot[only marks, mark=*, opacity=0.6 ,mark size=1.0, color=SpringGreen, error bars/.cd, 
    y fixed,
    y dir=both, 
    y explicit] table[x=Source_val,y=Source_MiniDN_AlexNet
]{./data/main/total_scatter.csv};
\addlegendentry{\scalebox{.8}{MiniDN, AlexNet}};
\end{axis}
\end{tikzpicture}

&

\pgfplotsset{every tick label/.append style={font=\tiny}}
\begin{tikzpicture}
\scriptsize
\begin{axis}[
    title style={yshift=-5pt,},
    title={Spearman's Corr. \green{\textbf{0.79}}},
    width=0.58\linewidth,
    height=0.41\linewidth,
    xmin=38.0,
    xmax=68.7,
    ymin=35.0,
    ymax=65.0,
    ytick={40,60},
    grid=both,
    xtick={40, 50, 60},
    xlabel style={yshift=5pt},
    title style={yshift=-3pt},
    legend pos=north west,
    legend style={cells={anchor=west}, font=\scriptsize, fill opacity=0.8, row sep=-2.5pt},
    ]
\addplot[only marks, mark=*, opacity=0.6 ,mark size=1.0, color=Green, error bars/.cd, 
    y fixed,
    y dir=both, 
    y explicit] table[x=ImgAug_val,y=ImgAug_MiniDN_ResNet
]{./data/main/total_scatter.csv};
\addplot[only marks, mark=*, opacity=0.6 ,mark size=1.0, color=SpringGreen, error bars/.cd, 
    y fixed,
    y dir=both, 
    y explicit] table[x=ImgAug_val,y=ImgAug_MiniDN_AlexNet
]{./data/main/total_scatter.csv};
\end{axis}
\end{tikzpicture}

\\

\pgfplotsset{every tick label/.append style={font=\tiny}}
\begin{tikzpicture}[define rgb/.code={\definecolor{mycolor}{RGB}{#1}}, rgb color/.style={define rgb={#1},mycolor}]
\scriptsize
\begin{axis}[
    title style={yshift=-5pt,},
    title={Spearman's Corr. \red{\textbf{-0.15}}},
    width=0.58\linewidth,
    height=0.41\linewidth,
    xmin=95.5,
    xmax=99.8,
    ymin=45.0,
    ymax=88.0,
    grid=both,
    xtick={97,99},
    ytick={40,60, 80},
    xlabel style={yshift=5pt},
    ylabel = {test-set accuracy},
    xlabel = {$V_S$: standard val. acc.},
    title style={yshift=-3pt},
    legend pos=north west,
    legend style={cells={anchor=west}, font =\scriptsize, fill opacity=0.6, row sep=-2.5pt,inner sep=1pt},
    ]
\addplot[only marks, mark=*, opacity=0.6 ,mark size=1.0, rgb color={72, 133, 250}, error bars/.cd, 
    y fixed,
    y dir=both, 
    y explicit] table[x=Source_val,y=Source_Digits_LeNet
]{./data/main/total_scatter.csv};
\addlegendentry{\scalebox{.8}{Digits, LeNet}};
\end{axis}
\end{tikzpicture}

&
\pgfplotsset{every tick label/.append style={font=\tiny}}
\begin{tikzpicture}[define rgb/.code={\definecolor{mycolor}{RGB}{#1}}, rgb color/.style={define rgb={#1},mycolor}]
\scriptsize
\begin{axis}[
    title style={yshift=-5pt,},
    title={Spearman's Corr. \green{\textbf{0.85}}},
    width=0.58\linewidth,
    height=0.41\linewidth,
    xmin=75.0,
    xmax=96.0,
    ymin=45.0,
    ymax=88.0,
    grid=both,
    xtick={80, 90},
    ytick={40,60, 80},
    ytick={40,60, 80},
    xlabel style={yshift=5pt},
    xlabel = {$V_A$: augmented val. acc.},
    title style={yshift=-3pt},
    legend pos=north west,
    legend style={cells={anchor=west}, font=\scriptsize, fill opacity=0.8, row sep=-2.5pt},
    ]
\addplot[only marks, mark=*, opacity=0.6 ,mark size=1.0, rgb color={72, 133, 250}, error bars/.cd, 
    y fixed,
    y dir=both, 
    y explicit] table[x=ImgAug_val,y=ImgAug_Digits_LeNet
]{./data/main/total_scatter.csv};
\end{axis}
\end{tikzpicture}
\end{tabular}
\vspace{-10pt}
\caption{
\textbf{Correlation between validation and test accuracy across the proposed variants}. Standard validation $V_S$ is performed on the validation set of the source domain, while the
proposed augmented validation $V_A$ uses images alternated by augmentations unseen during training.
Each point represents a different training-testing model variant.
}
\label{fig:all_datasets}
\vspace{-15pt}
\end{figure}

\subsection{Prior methods}
Publicly available implementations of SelfReg~\cite{SelfReg}, SagNet~\cite{SagNet}, L2D~\cite{Wlq+21}, and ACVC~\cite{acvc} are used for the comparison of our method with the state-of-the-art. These methods are also used to demonstrate the effectiveness of the proposed validation set on method selection. Implementation details regarding those methods are included in the Appendix.

\subsection{Results}

\begin{figure}[t]
\vspace{-5pt}
\begin{center}
\renewcommand{\arraystretch}{0.6}%
\begin{tabular}{@{\nsp}c@{\nsp}c@{\nsp}c}

\multicolumn{3}{c}{\small PACS - ResNet18} \\

\pgfplotsset{every tick label/.append style={font=\tiny}}
\begin{tikzpicture}[define rgb/.code={\definecolor{mycolor}{RGB}{#1}}, rgb color/.style={define rgb={#1},mycolor}]
\scriptsize
\begin{axis}[
    title = {Spearman's Corr. \textbf{\red{-0.20}}},
    width=0.54\linewidth,
  height=0.41\linewidth,
    xmin=96.5,
    xmax=100.2,
    ymin=29.0,
    ymax=72.0,
    grid=both,
    xtick={97,99},
    ytick={40,60, 80},
    title style = {yshift = -5pt},
    ylabel = {test-set accuracy},
    xticklabel style={yshift=-2pt},
    legend pos=north west,
    legend style={cells={anchor=east}, font =\fontsize{4}{3}\selectfont, fill opacity=0.7, row sep=-.5pt},
    ]

\addplot[only marks, mark=*, opacity=0.7 ,mark size=2, color=Red, error bars/.cd, 
    y fixed,
    y dir=both, 
    y explicit] table[x=Source_Val,y=Source_Val_Baseline
]{./data/main/corr_sota_pacs.csv};
\addlegendentry{Baseline};

\addplot[only marks, mark=*, opacity=0.7 ,mark size=2, color=Green, error bars/.cd, 
    y fixed,
    y dir=both, 
    y explicit] table[x=Source_Val,y=Source_Val_ACVC
]{./data/main/corr_sota_pacs.csv};
\addlegendentry{ACVC};

\addplot[only marks, mark=*, opacity=0.7 ,mark size=2, rgb color={72, 133, 250}, error bars/.cd, 
    y fixed,
    y dir=both, 
    y explicit] table[x=Source_Val,y=Source_Val_L2D
]{./data/main/corr_sota_pacs.csv};
\addlegendentry{L2D};

\addplot[only marks, mark=*, opacity=0.7 ,mark size=2, color=Orange, error bars/.cd, 
    y fixed,
    y dir=both, 
    y explicit] table[x=Source_Val,y=Source_Val_SelfReg
]{./data/main/corr_sota_pacs.csv};
\addlegendentry{SelfReg};

\addplot[only marks, mark=*, opacity=0.7 ,mark size=2, color=Dandelion, error bars/.cd, 
    y fixed,
    y dir=both, 
    y explicit] table[x=Source_Val,y=Source_Val_SagNet
]{./data/main/corr_sota_pacs.csv};
\addlegendentry{SagNet};

\addplot[only marks, mark=*, opacity=0.7 ,mark size=2, color=Plum, error bars/.cd, 
    y fixed,
    y dir=both, 
    y explicit] table[x=Source_Val,y=Source_Val_Ours_Full
]{./data/main/corr_sota_pacs.csv};
\addlegendentry{Ours};
\addplot [mark=star, color=white, mark size=2] coordinates {(100,43.3408) };
\end{axis}
\end{tikzpicture}

&

\pgfplotsset{every tick label/.append style={font=\tiny}}
\begin{tikzpicture}[define rgb/.code={\definecolor{mycolor}{RGB}{#1}}, rgb color/.style={define rgb={#1},mycolor}]
\scriptsize
\begin{axis}[
    title = {Spearman's Corr. \textbf{\green{0.74}}},
    width=0.54\linewidth,
  height=0.41\linewidth,
    xmin=64.0,
    xmax=92.0,
    ymin=29.0,
    ymax=72.0,
    grid=both,
    ytick={40,60, 80},
    title style = {yshift = -5pt},
    xticklabel style={yshift=-2pt},
    yticklabels=\empty,
    legend pos=south west,
    legend style={cells={anchor=east}, font =\footnotesize, fill opacity=0.7, row sep=-2.5pt},
    ]

\addplot[only marks, mark=*, opacity=0.7 ,mark size=2, color=Red, error bars/.cd, 
    y fixed,
    y dir=both, 
    y explicit] table[x=ImgAug_Val,y=ImgAug_Val_Baseline
]{./data/main/corr_sota_pacs.csv};

\addplot[only marks, mark=*, opacity=0.7 ,mark size=2, color=Green, error bars/.cd, 
    y fixed,
    y dir=both, 
    y explicit] table[x=ImgAug_Val,y=ImgAug_Val_ACVC
]{./data/main/corr_sota_pacs.csv};

\addplot[only marks, mark=*, opacity=0.7 ,mark size=2, rgb color={72, 133, 250}, error bars/.cd, 
    y fixed,
    y dir=both, 
    y explicit] table[x=ImgAug_Val,y=ImgAug_Val_L2D
]{./data/main/corr_sota_pacs.csv};

\addplot[only marks, mark=*, opacity=0.7 ,mark size=2, color=Orange, error bars/.cd, 
    y fixed,
    y dir=both, 
    y explicit] table[x=ImgAug_Val,y=ImgAug_Val_SelfReg
]{./data/main/corr_sota_pacs.csv};

\addplot[only marks, mark=*, opacity=0.7 ,mark size=2, color=Dandelion, error bars/.cd, 
    y fixed,
    y dir=both, 
    y explicit] table[x=ImgAug_Val,y=ImgAug_Val_SagNet
]{./data/main/corr_sota_pacs.csv};

\addplot[only marks, mark=*, opacity=0.7 ,mark size=2, color=Plum, error bars/.cd, 
    y fixed,
    y dir=both, 
    y explicit] table[x=ImgAug_Val,y=ImgAug_Val_Ours_Full
]{./data/main/corr_sota_pacs.csv};
\addplot [mark=star, color=white, mark size=2] coordinates {(89.01,65.54) };
\end{axis}
\end{tikzpicture}

&
\pgfplotsset{every tick label/.append style={font=\tiny}}
\begin{tikzpicture}
\scriptsize
    \begin{axis}[
        ybar=0pt,
        tickwidth = 0pt,
    width=0.33\linewidth,
  height=0.41\linewidth,
        symbolic x coords={$V_S$, $V_A$},
        xtick=data,
    ymin=29.0,
    ymax=72.0,
    grid=both,
    ytick={40,60, 80},
    enlarge x limits = 0.55,
    bar width=0.30cm,
    nodes near coords,
    yticklabels=\empty,
    tick label style={font=\tiny},
    nodes near coords style={font=\tiny}
    ]
    \addplot coordinates {($V_S$, 43.3) ($V_A$, 65.5)};
    \end{axis}
\end{tikzpicture}

\\

\multicolumn{3}{c}{\small Mini-DomainNet - ResNet18} \\

\pgfplotsset{every tick label/.append style={font=\tiny}}
\begin{tikzpicture}[define rgb/.code={\definecolor{mycolor}{RGB}{#1}}, rgb color/.style={define rgb={#1},mycolor}]
\scriptsize
\begin{axis}[
    title = {Spearman's Corr. \textbf{\red{-0.15}}},
    width=0.54\linewidth,
  height=0.41\linewidth,
    xmin=84.0,
    xmax=88.1,
    ymin=44.0,
    ymax=60.0,
    grid=both,
    xtick={85,87},
    ytick={45, 55},
    title style = {yshift = -5pt},
    ylabel = {test-set accuracy},
    xticklabel style={yshift=0pt},
    xlabel = {$V_S$ accuracy},
    legend pos=south west,
    legend style={cells={anchor=east}, font =\footnotesize, fill opacity=0.7, row sep=-2.5pt},
    ]

\addplot[only marks, mark=*, opacity=0.7 ,mark size=2, color=Red, error bars/.cd, 
    y fixed,
    y dir=both, 
    y explicit] table[x=Source_Val,y=Source_Val_Baseline
]{./data/main/corr_sota_minidn.csv};

\addplot[only marks, mark=*, opacity=0.7 ,mark size=2, color=Green, error bars/.cd, 
    y fixed,
    y dir=both, 
    y explicit] table[x=Source_Val,y=Source_Val_ACVC
]{./data/main/corr_sota_minidn.csv};

\addplot[only marks, mark=*, opacity=0.7 ,mark size=2, rgb color={72, 133, 250}, error bars/.cd, 
    y fixed,
    y dir=both, 
    y explicit] table[x=Source_Val,y=Source_Val_L2D
]{./data/main/corr_sota_minidn.csv};

\addplot[only marks, mark=*, opacity=0.7 ,mark size=2, color=Orange, error bars/.cd, 
    y fixed,
    y dir=both, 
    y explicit] table[x=Source_Val,y=Source_Val_SelfReg
]{./data/main/corr_sota_minidn.csv};

\addplot[only marks, mark=*, opacity=0.7 ,mark size=2, color=Dandelion, error bars/.cd, 
    y fixed,
    y dir=both, 
    y explicit] table[x=Source_Val,y=Source_Val_SagNet
]{./data/main/corr_sota_minidn.csv};

\addplot[only marks, mark=*, opacity=0.7 ,mark size=2, color=Plum, error bars/.cd, 
    y fixed,
    y dir=both, 
    y explicit] table[x=Source_Val,y=Source_Val_Ours_Full
]{./data/main/corr_sota_minidn.csv};

\addplot [mark=star, color=white, mark size=2] coordinates {(87.7,50.11) };

\end{axis}
\end{tikzpicture}

&
\pgfplotsset{every tick label/.append style={font=\tiny}}
\begin{tikzpicture}[define rgb/.code={\definecolor{mycolor}{RGB}{#1}}, rgb color/.style={define rgb={#1},mycolor}]
\scriptsize
\begin{axis}[
    title = {Spearman's Corr. \textbf{\green{0.83}}},
    width=0.54\linewidth,
  height=0.41\linewidth,
    xmin=47.0,
    xmax=67.0,
    ymin=44.0,
    ymax=60.0,
    grid=both,
    ytick={45,55},
    title style = {yshift = -5pt},
    xticklabel style={yshift=0pt},
    yticklabels=\empty,
    xlabel = {$V_A$ accuracy},
    legend pos=south east,
    legend style={cells={anchor=east}, font =\tiny, fill opacity=0.7, row sep=-2.5pt},
    ]

 \addplot[only marks, mark=*, opacity=0.7 ,mark size=2, color=Red, error bars/.cd, 
   y fixed,
   y dir=both, 
   y explicit] table[x=ImgAug_Val,y=ImgAug_Val_Baseline
 ]{./data/main/corr_sota_minidn.csv};

 \addplot[only marks, mark=*, opacity=0.7 ,mark size=2, color=Green, error bars/.cd, 
   y fixed,
   y dir=both, 
   y explicit] table[x=ImgAug_Val,y=ImgAug_Val_ACVC
 ]{./data/main/corr_sota_minidn.csv};

 \addplot[only marks, mark=*, opacity=0.7 ,mark size=2, rgb color={72, 133, 250}, error bars/.cd, 
   y fixed,
   y dir=both, 
   y explicit] table[x=ImgAug_Val,y=ImgAug_Val_L2D
 ]{./data/main/corr_sota_minidn.csv};

 \addplot[only marks, mark=*, opacity=0.7 ,mark size=2, color=Orange, error bars/.cd, 
   y fixed,
   y dir=both, 
   y explicit] table[x=ImgAug_Val,y=ImgAug_Val_SelfReg
 ]{./data/main/corr_sota_minidn.csv};

 \addplot[only marks, mark=*, opacity=0.7 ,mark size=2, color=Dandelion, error bars/.cd, 
   y fixed,
   y dir=both, 
   y explicit] table[x=ImgAug_Val,y=ImgAug_Val_SagNet]{./data/main/corr_sota_minidn.csv};
   
 \addplot[only marks, mark=*, opacity=0.7 ,mark size=2, color=Plum, error bars/.cd, 
   y fixed,
   y dir=both, 
   y explicit] table[x=ImgAug_Val,y=ImgAug_Val_Ours_Full
 ]{./data/main/corr_sota_minidn.csv};
\addplot [mark=star, color=white, mark size=2] coordinates {(64.1,57.09) };
\end{axis}
\end{tikzpicture}

&
\pgfplotsset{every tick label/.append style={font=\tiny}}
\begin{tikzpicture}
\scriptsize
    \begin{axis}[
    ybar,
    tickwidth = 0pt,
    width=0.33\linewidth,
    height=0.41\linewidth,
    symbolic x coords={$V_S$, $V_A$},
    xtick=data,
    ymin=44,
    ymax=60,
    ytick={45,55},
    grid=both,
    enlarge x limits = 0.55,
    bar width=0.30cm,
    nodes near coords,
    yticklabels=\empty,
    xlabel={Best model},
    tick label style={font=\tiny},
    nodes near coords style={font=\tiny},
    ]
    \addplot coordinates {($V_S$, 50.1) ($V_A$, 57.4)};
    \end{axis}
\end{tikzpicture}
\end{tabular}
\vspace{-15pt}
\caption{\textbf{Correlation between validation and test accuracy across literature methods} and our main variant $\hat{I}S \righttarrow I^{.75}S$ using standard $V_S$ and augmented $V_A$ validation set. The best model, according to each validation performance, is marked with a star. The test performance of the best model per validation set is summarized in the bar plot.
$V_A$ achieves significant test accuracy improvements of $22.2$ in PACS and $7.3$ in Mini-DomainNet.}
\label{fig:sota_scatter}
\vspace{-15pt}
\end{center}
\vspace{-15pt}
\end{figure}
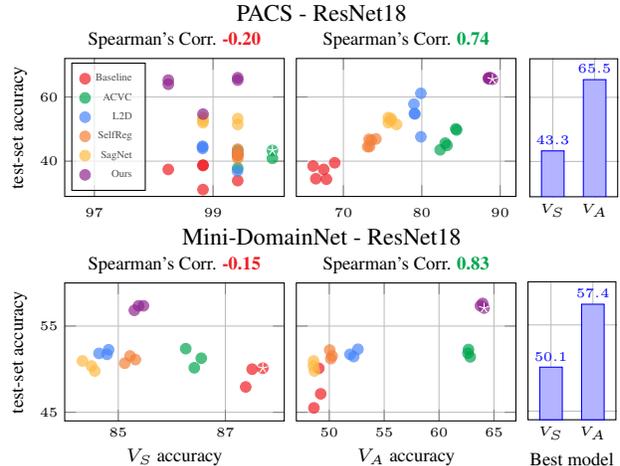

We perform two types of experiments: (i) to show the effectiveness of the proposed validation method, and (ii) to compare the proposed family of approaches with the state-of-the-art. Our main variant is $\hat{I}S \righttarrow I^{.75}S$, given that it is the top performing and is the most frequently selected by our validation method.

\paragraph{Correlation of validation and test performance.}
The models of the \emph{proposed variants} are evaluated on the standard validation set $V_S$, the proposed augmented validation $V_A$, and the test data. To visualize the reliability of predictions based on validation set performance, scatter plots comparing validation versus test performance are shown in Figure \ref{fig:all_datasets}. Performance on the proposed $V_A$ shows a much higher correlation with the test performance.
The performance saturation on $V_S$ for simple tasks like PACS is a major issue. 

A similar scatter plot for \emph{prior methods, the baseline, and our main variant} $\hat{I}S \righttarrow I^{.75}S$ is shown in Figure \ref{fig:sota_scatter}. Again, $V_A$ delivers better test performance prediction than $V_S$.
$V_S$ shows a weak negative correlation with the test performance and predicts that most existing methods, including ours, will not surpass the baseline.
As such, these methods would never be applied to the target domain in the real world, and we would never know that they work. $V_A$, on the other hand, achieves a strong positive correlation with the test set. The gains in test accuracy of using $V_A$ over $V_S$ are clearer in the bar plots (right). The performance of ACVC is over-estimated by $V_A$; we speculate this is because the ACVC training uses some of the extra augmentations included in $V_A$. Although we have not modified the ACVC method to fix this issue, the proposed TCV protocol is applicable. 

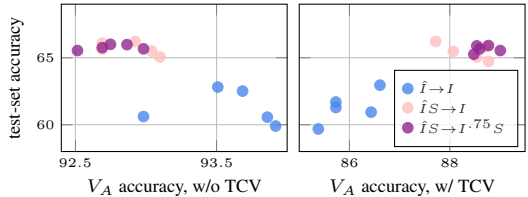
\begin{figure}[t]
\vspace{-5pt}
\centering
\begin{tabular}{@{\nsp}c@{\nsp}c}
\pgfplotsset{every tick label/.append style={font=\tiny}}
\begin{tikzpicture}[define rgb/.code={\definecolor{mycolor}{RGB}{#1}}, rgb color/.style={define rgb={#1},mycolor}]
\scriptsize
\begin{axis}[
    width=0.55\linewidth,
      height=0.425\linewidth,
    xmin=92.4,
    xmax=94.0,
    ymin=58.0,
    ymax=69.0,
    grid=both,
    xtick={92.5, 93.5},
    ytick={60, 65},
    ylabel = {test-set accuracy},
    xlabel = {$V_A$ accuracy, w/o TCV},
    legend pos=south east,
    legend style={cells={anchor=west}, font =\tiny, fill opacity=0.8, row sep=-2.5pt},
    ]

\addplot[only marks, mark=*, opacity=0.8 ,mark size=2.0, rgb color={72, 133, 237}, error bars/.cd, 
    y fixed,
    y dir=both, 
    y explicit] table[x=ImgAug_Val,y=ImgAug_Val_Augments_only
]{./data/main/2_fold_cv.csv};

\addplot[only marks, mark=*, opacity=0.8 ,mark size=2.0, color=pink, error bars/.cd, 
    y fixed,
    y dir=both, 
    y explicit] table[x=ImgAug_Val,y=ImgAug_Val_Ours_RGB_test
]{./data/main/2_fold_cv.csv};

\addplot[only marks, mark=*, opacity=0.8 ,mark size=2.0, color=Plum, error bars/.cd, 
    y fixed,
    y dir=both, 
    y explicit] table[x=ImgAug_Val,y=ImgAug_Val_Ours_Full
]{./data/main/2_fold_cv.csv};

\end{axis}
\end{tikzpicture}

&
\pgfplotsset{every tick label/.append style={font=\tiny}}
\begin{tikzpicture}[define rgb/.code={\definecolor{mycolor}{RGB}{#1}}, rgb color/.style={define rgb={#1},mycolor}]
\scriptsize
\begin{axis}[
    width=0.55\linewidth,
      height=0.425\linewidth,
    xmin=85.0,
    xmax=89.5,
    ymin=58.0,
    ymax=69.0,
    yticklabels=\empty,
    grid=both,
    xtick={86.0, 88.0},
    ytick={60, 65},
    xlabel = {$V_A$ accuracy, w/ TCV},
    legend pos=south east,
    legend style={cells={anchor=west}, font =\tiny, fill opacity=0.8, row sep=-2.5pt},
    ]

\addplot[only marks, mark=*, opacity=0.8 ,mark size=2.0, rgb color={72, 133, 237}, error bars/.cd, 
    y fixed,
    y dir=both, 
    y explicit] table[x=No_Cross_Val,y=No_Cross_Val_Augments_only
]{./data/main/2_fold_cv.csv};
\addlegendentry{$\hat{I} \righttarrow I$};

\addplot[only marks, mark=*, opacity=0.8 ,mark size=2.0, color=pink, error bars/.cd, 
    y fixed,
    y dir=both, 
    y explicit] table[x=No_Cross_Val,y=No_Cross_Val_Ours_RGB_test
]{./data/main/2_fold_cv.csv};
\addlegendentry{$\hat{I}S \righttarrow I$};

\addplot[only marks, mark=*, opacity=0.8 ,mark size=2.0, color=Plum, error bars/.cd, 
    y fixed,
    y dir=both, 
    y explicit] table[x=No_Cross_Val,y=No_Cross_Val_Ours_Full
]{./data/main/2_fold_cv.csv};
\addlegendentry{$\hat{I}S\righttarrow I^{.75}S$};

\end{axis}
\end{tikzpicture}
\end{tabular}
\vspace{-15pt}
\caption{\textbf{Correlation between validation and test accuracy with or without TCV} when using $V_A$ as the validation set. The experiment is conducted on PACS using ResNet-18.
\label{fig:abla_cros_val}
\vspace{-15pt}
}
\end{figure}

We perform an experiment to highlight the importance of the \emph{TCV protocol}, \ie, avoiding training and validating on the same pool of augmentations. Figure~\ref{fig:abla_cros_val} shows the correlation of the $V_A$ validation with the test performance, with and without TCV. Skipping TCV leads to performance over-estimation for methods that use the same extra augmentations for both training and validation. This is a similar effect to what is observed with validating on the training distribution in Figure~\ref{fig:all_datasets}.

\paragraph{Performance gains by better validation.}
We conduct an experiment for \emph{method selection} across seven dataset-backbone configurations. The validation process is responsible for tuning the learning rate and selecting the best method out of our variants. More than $4,500$ models are trained for this experiment, which effectively increases to more than $15,000$ when combined with our testing variants. The results are summarised in Figure~\ref{fig:radar}. The test accuracy of the models selected by $V_A$ significantly surpasses the performance of the ones chosen by $V_S$ for every benchmark. $V_A$ achieves more than $98\%$ of the oracle performance on average. 

\begin{figure}[t]
\vspace{-5pt}
\centering
\def\n{8}   
\def\N{5}   
\def\minvala{8.0}
\def\minvalb{8.0}
\begin{tikzpicture}[scale=0.28]

\foreach \x in{1,...,\n}{%
    \draw[thin,lightgray] (-0.5,\x)--(10,\x);
}

\draw(5,10)node[text width=4cm,align=center]{\footnotesize tune learning rate };
\draw[thin,lightgray] (0,0.5)--(0,8.5);
\pgfmathparse{0.1*\minvala}
\edef\xcoord{\pgfmathprintnumber[fixed, precision=1]{\pgfmathresult}}
\pgfmathparse{0.1*\minvalb}

\draw(0,9)node[text width=0.5cm,align=center]{\tiny \xcoord}; 
\draw[thin,lightgray] (5,0.5)--(5,8.5);
\pgfmathparse{0.1*(\minvala+10)/2}
\edef\xcoord{\pgfmathprintnumber[fixed, precision=2]{\pgfmathresult}}
\draw(5,9)node[text width=0.5cm,align=center]{\tiny \xcoord}; 
\draw[thin,lightgray] (10,0.5)--(10,8.5);
\draw(10,9)node[text width=0.5cm,align=center]{\tiny 1.0};

\draw(0,8)node[text width=4cm,align=left]{\fontsize{7}{8}\selectfont PACS-RN18}; 
\draw(0,7)node[text width=4cm,align=left]{\fontsize{7}{8}\selectfont PACS-ViT-S};  
\draw(0,6)node[text width=4cm,align=left]{\fontsize{7}{8}\selectfont MiniDN-AlexNet};
\draw(0,5)node[text width=4cm,align=left]{\fontsize{7}{8}\selectfont MiniDN-RN18};
\draw(0,4)node[text width=4cm,align=left]{\fontsize{7}{8}\selectfont Digits-LeNet};
\draw(0,3)node[text width=4cm,align=left]{\fontsize{7}{8}\selectfont Camelyon-RN50};
\draw(0,2)node[text width=4cm,align=left]{\fontsize{7}{8}\selectfont NICO++-RN18};
\draw(0,1)node[text width=4cm,align=left]{\fontsize{7}{8}\selectfont Average};

\pgfmathparse{(9.239-\minvala)*10/(10 - \minvala)}
\draw[fill=Orange, Orange, opacity=0.7] (\pgfmathresult,8) circle(7pt);
\pgfmathparse{(9.697-\minvala)*10/(10 - \minvala)}
\draw[fill=Orange, Orange, opacity=0.7] (\pgfmathresult,7) circle(7pt);
\pgfmathparse{(9.927-\minvala)*10/(10 - \minvala)}
\draw[fill=Orange, Orange, opacity=0.7] (\pgfmathresult,6) circle(7pt);
\pgfmathparse{(9.909-\minvala)*10/(10 - \minvala)}
\draw[fill=Orange, Orange, opacity=0.7] (\pgfmathresult,5) circle(7pt);
\pgfmathparse{(9.660-\minvala)*10/(10 - \minvala)}
\draw[fill=Orange, Orange, opacity=0.7] (\pgfmathresult,4) circle(7pt);
\pgfmathparse{(9.926-\minvala)*10/(10 - \minvala)}
\draw[fill=Orange, Orange, opacity=0.7] (\pgfmathresult,3) circle(7pt);
\pgfmathparse{(9.807-\minvala)*10/(10 - \minvala)}
\draw[fill=Orange, Orange, opacity=0.7] (\pgfmathresult,2) circle(7pt);
\pgfmathparse{((9.735-\minvala)*10/(10 - \minvala)}
\draw[fill=Orange, Orange, opacity=0.7] (\pgfmathresult,1) circle(7pt);

\pgfmathparse{((9.651-\minvala)*10/(10 - \minvala)}
\draw[fill=RoyalBlue, RoyalBlue, opacity=0.7] (\pgfmathresult,8) circle(7pt);
\pgfmathparse{((9.947-\minvala)*10/(10 - \minvala)}
\draw[fill=RoyalBlue, RoyalBlue, opacity=0.7] (\pgfmathresult,7) circle(7pt);
\pgfmathparse{((9.938-\minvala)*10/(10 - \minvala)}
\draw[fill=RoyalBlue, RoyalBlue, opacity=0.7] (\pgfmathresult,6) circle(7pt);
\pgfmathparse{((9.868-\minvala)*10/(10 - \minvala)}
\draw[fill=RoyalBlue, RoyalBlue, opacity=0.7] (\pgfmathresult,5) circle(7pt);
\pgfmathparse{((9.965-\minvala)*10/(10 - \minvala)}
\draw[fill=RoyalBlue, RoyalBlue, opacity=0.7] (\pgfmathresult,4) circle(7pt);
\pgfmathparse{((9.984-\minvala)*10/(10 - \minvala)}
\draw[fill=RoyalBlue, RoyalBlue, opacity=0.7] (\pgfmathresult,3) circle(7pt);
\pgfmathparse{((9.755-\minvala)*10/(10 - \minvala)}
\draw[fill=RoyalBlue, RoyalBlue, opacity=0.7] (\pgfmathresult,2) circle(7pt);
\pgfmathparse{((9.894-\minvala)*10/(10 - \minvala)}
\draw[fill=RoyalBlue, RoyalBlue, opacity=0.7] (\pgfmathresult,1) circle(7pt);

\foreach \x in{1,...,\n}{%
    \draw[thin,lightgray] (11.5,\x)--(22,\x);
}

\draw(17,10)node[text width=4cm,align=center]{\footnotesize tune loss weight $\lambda$};

\draw[thin,lightgray] (12,0.5)--(12,8.5);
\pgfmathparse{0.1*\minvalb}
\edef\xcoord{\pgfmathprintnumber[fixed, precision=1]{\pgfmathresult}}
\draw(12,9)node[text width=0.5cm,align=center]{\tiny \xcoord}; 
\draw[thin,lightgray] (17,0.5)--(17,8.5);
\pgfmathparse{0.1*(\minvalb+10)/2}
\edef\xcoord{\pgfmathprintnumber[fixed, precision=2]{\pgfmathresult}}
\draw(17,9)node[text width=0.5cm,align=center]{\tiny \xcoord}; 
\draw[thin,lightgray] (22,0.5)--(22,8.5);
\draw(22,9)node[text width=0.5cm,align=center]{\tiny 1.0}; 

\pgfmathparse{((8.119-\minvalb)*10/(10 - \minvalb)}
\draw[fill=Orange, Orange, opacity=0.7] (12+\pgfmathresult,8) circle(7pt);
\pgfmathparse{((9.889-\minvalb)*10/(10 - \minvalb)}
\draw[fill=Orange, Orange, opacity=0.7] (12+\pgfmathresult,7) circle(7pt);
\pgfmathparse{((9.485-\minvalb)*10/(10 - \minvalb)}
\draw[fill=Orange, Orange, opacity=0.7] (12+\pgfmathresult,6) circle(7pt);
\pgfmathparse{((9.128-\minvalb)*10/(10 - \minvalb)}
\draw[fill=Orange, Orange, opacity=0.7] (12+\pgfmathresult,5) circle(7pt);
\pgfmathparse{((9.849-\minvalb)*10/(10 - \minvalb)}
\draw[fill=Orange, Orange, opacity=0.7] (12+\pgfmathresult,4) circle(7pt);
\pgfmathparse{((9.960-\minvalb)*10/(10 - \minvalb)}
\draw[fill=Orange, Orange, opacity=0.7] (12+\pgfmathresult,3) circle(7pt);
\pgfmathparse{((9.748-\minvalb)*10/(10 - \minvalb)}
\draw[fill=Orange, Orange, opacity=0.7] (12+\pgfmathresult,2) circle(7pt);
\pgfmathparse{((9.501-\minvalb)*10/(10 - \minvalb)}
\draw[fill=Orange, Orange, opacity=0.7] (12+\pgfmathresult,1) circle(7pt);

\pgfmathparse{((9.521-\minvalb)*10/(10 - \minvalb)}
\draw[fill=RoyalBlue, RoyalBlue, opacity=0.7] (12+\pgfmathresult,8) circle(7pt);
\pgfmathparse{((9.916-\minvalb)*10/(10 - \minvalb)}
\draw[fill=RoyalBlue, RoyalBlue, opacity=0.7] (12+\pgfmathresult,7) circle(7pt);
\pgfmathparse{((9.891-\minvalb)*10/(10 - \minvalb)}
\draw[fill=RoyalBlue, RoyalBlue, opacity=0.7] (12+\pgfmathresult,6) circle(7pt);
\pgfmathparse{((9.962-\minvalb)*10/(10 - \minvalb)}
\draw[fill=RoyalBlue, RoyalBlue, opacity=0.7] (12+\pgfmathresult,5) circle(7pt);
\pgfmathparse{((9.887-\minvalb)*10/(10 - \minvalb)}
\draw[fill=RoyalBlue, RoyalBlue, opacity=0.7] (12+\pgfmathresult,4) circle(7pt);
\pgfmathparse{((9.960-\minvalb)*10/(10 - \minvalb)}
\draw[fill=RoyalBlue, RoyalBlue, opacity=0.7] (12+\pgfmathresult,3) circle(7pt);
\pgfmathparse{((9.955-\minvalb)*10/(10 - \minvalb)}
\draw[fill=RoyalBlue, RoyalBlue, opacity=0.7] (12+\pgfmathresult,2) circle(7pt);
\pgfmathparse{((9.871-\minvalb)*10/(10 - \minvalb)}
\draw[fill=RoyalBlue, RoyalBlue, opacity=0.7] (12+\pgfmathresult,1) circle(7pt);

\draw[fill=RoyalBlue, RoyalBlue, opacity=0.7] (-5.6,-1) circle(7pt);
\node at (1,-1) {Proposed $V_A$ Validation};
\draw[fill=Orange, Orange, opacity=0.7] (8.5,-1) circle(7pt);
\node at (15,-1) {Standard $V_S$ Validation};

\end{tikzpicture}
\vspace{-10pt}
\caption{\textbf{Learning rate and loss weight tuning per validation.} We fix the training-testing variant to $IS\righttarrow I^{.75}S$ to evaluate the performance of $V_A$ and $V_S$ validation restricted only to hyper-parameter tuning. Performance is normalized by the performance of the oracle; 1 means equivalent to oracle performance. 
\label{fig:dot_plots}
\vspace{-5pt}
}
\end{figure}
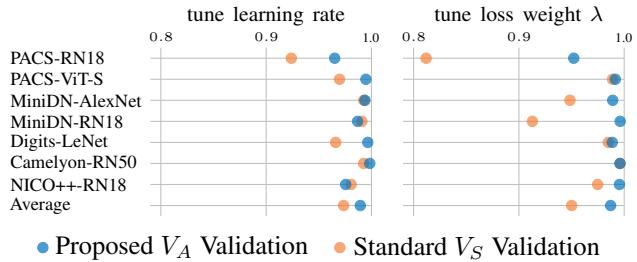

We perform a hyper-parameter tuning experiment, independently tuning the learning rate and the loss weight $\lambda$. We report the performance of the selected models by the two validation processes in Figure~\ref{fig:dot_plots}. The average gain in absolute (relative) test accuracy of $V_A$ compared to $V_S$ is $1.0$ (1.6\%) for the learning rate and $2.3$ (3.9\%) for the weight of the loss. Although $V_A$ selects models that perform better in both tuning processes, a higher gain is observed when tuning $\lambda$, a hyper-parameter related to domain generalization. On the other hand, the impact of a hyper-parameter related to the learning process, \ie, learning rate, is also expected to be reflected on the validation set from the training distribution. Even in this favorable scenario for $V_S$, there is no evidence to support using $V_S$ over the proposed $V_A$.

\paragraph{Impact of components, extra augmentations, and shape usage in training/testing.}
To study and show the impact of various components of the proposed methods, nine variants are compared in Table~\ref{tab:training_ablations}. The method denoted by $S_{sob}$ uses a non-binarized Sobel edge map instead of BTE, and the method $\hat{I}_{+BTE}$ uses BTE as an additional extra augmentation applied randomly to some images, as opposed to standard input processing applied to all images.

Including shape information during training (var. 2 and 3) gives a noticeable boost  in performance across all domains compared to the baseline (var. 1) despite not using shape during testing. Since both images and shapes are processed by the same classifier during training, this is a way to make the network focus on shapes even when the test-time input is only an image.
Shape in the form of BTEs (var. 3) contributes much more than the continuous edge maps from the Sobel operator (var. 2), especially in target domains that lack texture information. Additionally, using shape during test time gives a further boost (var. 4).
Adding the extra augmentations in training has a positive impact (var. 5 vs. 1, 7 vs. 3, and 8 vs. 4).
Using BTE as yet another augmentation type increases the performance (var. 6 vs. 5) but not in the extend of using it on every training image (var. 7 vs. 6) or on test images (var. 8 vs. 6).
Lastly, we evaluate the ensemble of two separate networks, one for images and one for shapes. We observe that the single network approach is not only more efficient but also significantly better (var. 9 vs. 4). We conclude that it is not only the ensemble of the two predictions that improves the test accuracy but mainly the bridging of the two domains during training. 

\paragraph{Beyond shape-oriented domains.}  PACS and Mini-DomainNet include domains where the shape is preserved, but the texture varies. NICO++ differs because it consists of natural images of objects in varying contexts and environments. We use NICO++ to highlight the generality of the proposed recognition method. Due to the increased experimental cost, we only include the variants $I\righttarrow I$, $IS\righttarrow I^{.75}S$, and $\hat{I}S\righttarrow I^{.75}S$. The baseline $I\righttarrow I$ achieves a $23.8$ test accuracy, while including BTEs achieve a $2.2$ increase over the baseline, confirming that BTEs not only help the network learn a shape-oriented representation but also enhance robustness against spurious correlations from textures. $\hat{I}S\righttarrow I^{.75}S$ achieves an additional $1.3$ increase.

\begin{table}[t]
\centering
\fontsize{8}{6}\selectfont
\def\arraystretch{1.2}
\begin{tabular}{@{\msp}l@{\lsp}l@{\lsp}l@{\msp}l@{\msp}l@{\msp}}
\hline
Variant                                                       & Avg.         & Art   & Cart. & Sketch \\ \hline \hline
1: $I\righttarrow I$   (baseline)                   & 43.46        & 54.30 & 37.27   & 38.80  \\
2: $IS_{sob}\righttarrow I$                  & 51.75        & 58.35 & 41.19   & 55.71  \\
3: $IS\righttarrow I$                          & 57.89        & 58.47 & 48.21   & 67.00  \\
4: $IS\righttarrow I^{.75}S$                   & 59.02        & 59.10 & 49.54   & 68.41  \\
 5:  $\hat{I}\righttarrow I$            &  58.15        &  63.10 &  48.46   &  62.91  \\
 6:  $\hat{I}_{+BTE}\righttarrow I$      &  59.31        &  63.38 &  48.74   &  65.83  \\
7: $\hat{I}S\righttarrow I$                    & 60.50        & 63.23 & 50.62   & 67.66  \\
8: $\hat{I}S\righttarrow I^{.75}S$             & 60.97        & 63.62 & 50.80   & 68.50  \\
9: $IS_{\times 2}\righttarrow I^{.75}S_{\times 2}$      & 53.34        & 57.52 & 44.79   & 57.70 \\ \hline
\end{tabular}

\caption{\textbf{Comparison of variants.} Performance comparison among our variants on the PACS dataset with AlexNet. The value of the shape information and the extra augmentations is showcased. We report accuracy per target domain and their average.
\label{tab:training_ablations}
\vspace{-15pt}
}
\end{table}

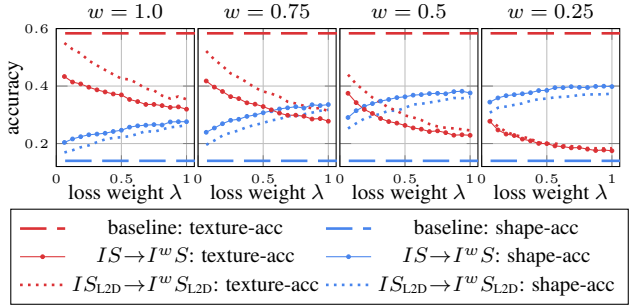
\begin{figure}[t]
\vspace{-5pt}
\centering
\definecolor{myblue}{RGB}{72, 133, 237}
\pgfplotsset{every tick label/.append style={font=\tiny}}
\begin{tikzpicture}[define rgb/.code={\definecolor{mycolor}{RGB}{#1}}, rgb color/.style={define rgb={#1},mycolor}]
\footnotesize
\begin{axis}[
    title={$w=1.0$},
    title style = {yshift=-5pt},
    width=.41\linewidth,
    height=.41\linewidth,
    name=plot1,
    ymin=0.12,
    ymax=0.6,
    xmin=0,xmax=1.06,
    ylabel style = {yshift = -5pt},
    ylabel = {accuracy},
    xlabel = {loss weight $\lambda$},
    xlabel style={yshift=5pt},
    grid=both,
    legend pos=south east,
    legend style={cells={anchor=west}, font =\scriptsize, fill opacity=0.7, row sep=-2.9pt, xshift=-26ex, yshift=5ex, inner sep=1pt},
    ]

\addplot[dotted, mark=none,line width=1pt,rgb color={72, 133, 237}, error bars/.cd, 
    y fixed,
    y dir=both, 
    y explicit] table[x=method_loss,y=Shape_augnet
]{./data/main/shape_vs_texture.csv};

\addplot[dotted, mark=none,line width=1pt,rgb color={219,  50,  54}, error bars/.cd, 
    y fixed,
    y dir=both, 
    y explicit] table[x=method_loss,y=Texture_augnet
]{./data/main/shape_vs_texture.csv};

\addplot[mark=*,mark size=0.7,rgb color={72, 133, 237}, error bars/.cd, 
    y fixed,
    y dir=both, 
    y explicit] table[x=method_loss,y=Shape_100
]{./data/main/shape_vs_texture.csv};

\addplot[mark=*,mark size=0.7,rgb color={219,  50,  54}, error bars/.cd, 
    y fixed,
    y dir=both, 
    y explicit] table[x=method_loss,y=Texture_100
]{./data/main/shape_vs_texture.csv};

\addplot[mark=none,mark size=0.7,dash pattern=on 10 off 4,line width=1pt, rgb color={72, 133, 237}, error bars/.cd] coordinates {(-0.5,0.13984) (1.5,0.13984)};

\addplot[mark=none,mark size=0.7,line width=1pt, rgb color={219,  50,  54}, dash pattern=on 10 off 4, error bars/.cd] coordinates {(-0.5,0.5835) (1.5,0.5835)};

\end{axis}

\begin{axis}[
    title={$w=0.75$},
    title style = {yshift=-5pt},
    width=.41\linewidth,
    height=.41\linewidth,
    name=plot2,
    at={(plot1.right of east)},
    anchor=west,
    ymin=0.12,
    ymax=0.6,
    xmin=0,xmax=1.06,
    yticklabels=\empty,
    xlabel = {loss weight $\lambda$},
    xlabel style={yshift=5pt},
    grid=both,
    legend pos=south east,
    legend style={cells={anchor=west}, font =\footnotesize, fill opacity=0.7, row sep=-2.9pt, yshift=22ex},
    ]
\addplot[mark=none, dash pattern=on 10 off 4, mark size=0.7,line width=1pt, rgb color={72, 133, 237}, error bars/.cd] coordinates {(-0.5,0.13984) (1.5,0.13984)};
\addplot[mark=none, dash pattern=on 10 off 4, mark size=0.7,line width=1pt, rgb color={219,  50,  54}, error bars/.cd] coordinates {(-0.5,0.5835) (1.5,0.5835)};
\addplot[mark=*,mark size=0.7,rgb color={72, 133, 237}, error bars/.cd, 
    y fixed,
    y dir=both, 
    y explicit] table[x=method_loss,y=Shape_75
]{./data/main/shape_vs_texture.csv};
\addplot[mark=*,mark size=0.7,rgb color={219,  50,  54}, error bars/.cd, 
    y fixed,
    y dir=both, 
    y explicit] table[x=method_loss,y=Texture_75
]{./data/main/shape_vs_texture.csv};
\addplot[dotted, mark=none,line width=1pt,rgb color={72, 133, 237}, error bars/.cd, 
    y fixed,
    y dir=both, 
    y explicit] table[x=method_loss,y=Shape_augnet_75
]{./data/main/shape_vs_texture.csv};
\addplot[dotted, mark=none,line width=1pt,rgb color={219,  50,  54}, error bars/.cd, 
    y fixed,
    y dir=both, 
    y explicit] table[x=method_loss,y=Texture_augnet_75
]{./data/main/shape_vs_texture.csv};
\end{axis}

\begin{axis}[
    title={$w=0.5$},
    title style = {yshift=-5pt},
    width=.41\linewidth,
    height=.41\linewidth,
    name=plot3,
    at={(plot2.right of east)},
    anchor=west,
    ymin=0.12,
    ymax=0.6,
    xmin=0,xmax=1.06,
    yticklabels=\empty,
    xlabel = {loss weight $\lambda$},
    xlabel style={yshift=5pt},
    grid=both,
    legend pos=south east,
    legend style={cells={anchor=west}, font =\footnotesize, fill opacity=0.7, row sep=-2.9pt, yshift=22ex},
    ]

\addplot[mark=*,mark size=0.7,rgb color={72, 133, 237}, error bars/.cd, 
    y fixed,
    y dir=both, 
    y explicit] table[x=method_loss,y=Shape_50
]{./data/main/shape_vs_texture.csv};
\addplot[mark=*,mark size=0.7,rgb color={219,  50,  54}, error bars/.cd, 
    y fixed,
    y dir=both, 
    y explicit] table[x=method_loss,y=Texture_50
]{./data/main/shape_vs_texture.csv};
\addplot[mark=none,mark size=0.7,dash pattern=on 10 off 4, line width=1pt, rgb color={219,  50,  54}, error bars/.cd] coordinates {(-0.5,0.5835) (1.5,0.5835)};
\addplot[mark=none,mark size=0.7,dash pattern=on 10 off 4,line width=1pt, rgb color={72, 133, 237}, error bars/.cd] coordinates {(-0.5,0.13984) (1.5,0.13984)};
\addplot[dotted, mark=none,line width=1pt,rgb color={72, 133, 237}, error bars/.cd, 
    y fixed,
    y dir=both, 
    y explicit] table[x=method_loss,y=Shape_augnet_50
]{./data/main/shape_vs_texture.csv};
\addplot[dotted, mark=none,line width=1pt,rgb color={219,  50,  54}, error bars/.cd, 
    y fixed,
    y dir=both, 
    y explicit] table[x=method_loss,y=Texture_augnet_50
]{./data/main/shape_vs_texture.csv};

\end{axis}

\begin{axis}[
    title={$w=0.25$},
    title style = {yshift=-5pt},
    width=.41\linewidth,
    height=.41\linewidth,
    name=plot4,
    at={(plot3.right of east)},
    ymin=0.12,
    ymax=0.6,
    xmin=0,xmax=1.06,
    yticklabels=\empty,
    xlabel = {loss weight $\lambda$},
    xlabel style={yshift=5pt},
    grid=both,
    anchor=west,
    legend to name=named,
    legend columns=2,
    legend style={/tikz/every even column/.append style={column sep=0cm, font =\footnotesize}},
    ]

\addplot[mark=none,mark size=0.7, dash pattern=on 10 off 4,line width=1pt, rgb color={219,  50,  54}, error bars/.cd] coordinates {(-0.5,0.5835) (1.5,0.5835)};
\addlegendentry{baseline: texture-acc};

\addplot[mark=none,mark size=0.7,dash pattern=on 10 off 4,line width=1pt, rgb color={72, 133, 237}, error bars/.cd] coordinates {(-0.5,0.13984) (1.5,0.13984)};
\addlegendentry{baseline: shape-acc};

\addplot[mark=*,mark size=0.7, rgb color={219,  50,  54}, error bars/.cd, 
    y fixed,
    y dir=both, 
    y explicit] table[x=method_loss,y=Texture_25
]{./data/main/shape_vs_texture.csv};
\addlegendentry{$IS\righttarrow I^{w}S$: texture-acc};

\addplot[mark=*,mark size=0.7, rgb color={72, 133, 237}, error bars/.cd, 
    y fixed,
    y dir=both, 
    y explicit] table[x=method_loss,y=Shape_25
]{./data/main/shape_vs_texture.csv};
\addlegendentry{$IS\righttarrow I^{w}S$: shape-acc};

\addplot[dotted, mark=none,line width=1pt,rgb color={219,  50,  54}, error bars/.cd, 
    y fixed,
    y dir=both, 
    y explicit] table[x=method_loss,y=Texture_augnet_25
]{./data/main/shape_vs_texture.csv};
\addlegendentry{$I\SL2D\righttarrow I^w\SL2D$: texture-acc};

\addplot[dotted, mark=none,line width=1pt,rgb color={72, 133, 237}, error bars/.cd, 
    y fixed,
    y dir=both, 
    y explicit] table[x=method_loss,y=Shape_augnet_25
]{./data/main/shape_vs_texture.csv};
\addlegendentry{$I\SL2D\righttarrow I^w\SL2D$:   shape-acc}

\end{axis}
\node at (3.5,-1.2) {\pgfplotslegendfromname{named}};
\end{tikzpicture}
\vspace{-15pt}
\caption{\textbf{Shape-texture bias:} The accuracy of different variants is evaluated on the cue conflict stimuli test set based on shape or test labels. Loss weight $\lambda$ and exponent $w$ are the ways to control the influence of shape during training and testing, respectively, with $\lambda=1$ and $w=0$ taking shape into account the most. A different model is trained for the various values of $\lambda$.
\label{fig:shape_vs_texture}
\vspace{-15pt}
}
\end{figure}

\paragraph{Shape-texture bias.} 
The inductive biases~\cite{grm+19} of our approach are analyzed by using the \emph{16-class-ImageNet} \cite{gtr+18} as the source domain and the \emph{cue conflict stimuli} images \cite{grm+19} as the target domain. 
Each stimuli image is the product of blending an image's texture with another's shape through style transfer. Therefore, each image has two labels, one based on shape and one based on texture. Although the 16-class-ImageNet consists of $213,555$ images, we sub-sample to $500$ images per class.
An ImageNet pre-trained ResNet18 backbone is trained with the $IS$ training variant for varying values of the loss weight $\lambda$. 
We compare our variant $IS\righttarrow I^w S$ with a stylization-based approach, a popular shape bias technique. In particular, the style-complement component of L2D~\cite{Wlq+21} is used as a shape component, replacing the BTE. This component is randomized during training and fixed during testing. We refer to this method as $I\SL2D\righttarrow I^w \SL2D$, and the results are summarized in Figure \ref{fig:shape_vs_texture}, where accuracy is evaluated either with the shape (shape-acc.) or the texture labels (texture-acc.).

The results demonstrate the significant impact of both shape-controlling hyper-parameters, namely $\lambda$ during training and $w$ during testing, on shape bias. The comparison shows a higher ability of BTE to adapt to shape cues compared to the stylization of L2D. This seems crucial for domain generalization since humans achieve an accuracy of $95.9$ regarding the shape labels~\cite{grm+19}.

\begin{table}[t]
\centering
\label{tab:combined_sota}

\begin{minipage}{\columnwidth}
\begin{subtable}{\columnwidth}
\centering
\fontsize{6}{6}\selectfont
\def\arraystretch{1.15}
\begin{tabular}{@{\xssp}l@{\xtssp}l@{\xtssp}l@{\xtssp}l@{\xtssp} l@{\xtssp}l@{\xtssp}l@{\xtssp} }
\hline
Method & SVHN &  MNIST-M & SYN & USPS & Avg. \\ \hline \hline
ERM \cite{ProRandConv} & 32.52 & 54.92 & 42.34 & 78.21 & 52.00 \\
RandConv \cite{RandConv} & 62.07 & \underline{87.89} & 63.90 & 84.39 & 74.56 \\
L2D \cite{Wlq+21} & 62.86 & 87.30 & 63.72 & 83.97 & 74.46 \\
MetaCNN \cite{wsz22} & 66.50 & \textbf{88.27} & 70.66 & 89.64 & 78.76 \\
MCL \cite{MCL} & \textbf{69.94} & 78.34 & 78.47 & 88.54 & 78.82 \\
ProRandC \cite{ProRandConv} &  \underline{69.67} & 82.30 & \textbf{79.77} & 93.67 & \underline{81.35} \\
CADA \cite{CADA} &  67.27 & 78.66 & 79.34 & \underline{96.96} & 80.56 \\ 
ABA$_{3l+RC}$ \cite{ABA} & 56.87 & 80.08 & 73.40 & 96.55 & 76.72 \\ \hline
\tiny \textls[-140]{\checkmark $\hat{I}S\righttarrow I^{.75}S$} (Ours)  & 67.82 \textls[-140]{ \scalebox{0.8}{$\pm$ 1.0}} & 84.28 \textls[-140]{\scalebox{0.8}{$\pm$ 0.4}} & \underline{79.64} \textls[-140]{\scalebox{0.8}{$\pm$ 1.0}} & \textbf{98.68} \textls[-140]{\scalebox{0.8}{$\pm$ 0.1}} & \textbf{82.61} \textls[-140]{\scalebox{0.8}{$\pm$ 0.5}} \\

\hline
\end{tabular}

\caption{\textbf{Digits with LeNet.}
\label{tab:Digits}} 
\centering
\fontsize{6}{6}\selectfont
\def\arraystretch{1.15}
\begin{tabular}{l@{\mllsp}l@{\mllsp}l@{\mllsp}l@{\mllsp} l@{\xtssp} }
\hline
Method & Clipart &  Painting & Sketch &  Avg. \\ \hline \hline
ERM$^\dag$ & 50.53 & 53.86 & 38.36 & 47.58  \\
SagNet$^\dag$ \cite{SagNet} & 49.63 & 55.66 & 45.82 & 50.37 \\
ACVC$^\dag$ \cite{acvc} & 53.81 & \underline{56.93} & 43.17 & 51.27 \\
SelfReg$^\dag$ \cite{SelfReg} & 52.96 & 53.76 & \underline{48.25} & 51.66 \\
L2D$^\dag$ \cite{Wlq+21} & \underline{54.95} & 55.38 & 45.15 & \underline{51.83} \\
 \hline
\tiny \textls[-140]{\checkmark $\hat{I}S\righttarrow I^{.75}S$} (Ours)  & \textbf{55.55} \textls[-140]{ \scalebox{0.8}{$\pm$ 0.3}} &  \textbf{59.00} \textls[-140]{\scalebox{0.8}{$\pm$ 0.5}} &  \textbf{57.51} \textls[-140]{\scalebox{0.8}{$\pm$ 1.2}} &  \textbf{57.35} \textls[-140]{\scalebox{0.8}{$\pm$ 0.3}} \\

\hline
\end{tabular}

\caption{\textbf{Mini-DomainNet with ResNet18.}
\vspace{-10pt}
\label{tab:miniDN_sota}}
\centering
\begin{center}
\fontsize{6}{6}\selectfont

\def\arraystretch{1.25}
\begin{tabular}{l@{\mllsp}l@{\mllsp}l@{\mllsp}l@{\mllsp} l@{\xtssp} }
\hline
\multicolumn{5}{c}{Alexnet}  \\
 \hline
Method  & P \hspace{-3pt}$\rightarrow{}$\hspace{-4pt} A & P \hspace{-3pt}$\rightarrow{}$\hspace{-4pt} C & P \hspace{-3pt}$\rightarrow{}$\hspace{-4pt} S &  Avg. \\ \hline \hline
ERM & 54.43 & 42.74 & 42.02 & 46.39 \\
L2D \cite{Wlq+21} & \underline{56.26} & \underline{51.04} & 58.42 & 55.24 \\
MetaCNN \cite{wsz22}& 54.05 & \textbf{53.58} & \underline{63.88} & \underline{57.17}  \\ \hline
\tiny \textls[-140]{\checkmark $\hat{I}S\righttarrow I^{.75}S$} (Ours)  & \textbf{63.62}  \textls[-140]{\scalebox{0.8}{$\pm$ 0.4}} & 50.80  \textls[-140]{\scalebox{0.8}{$\pm$ 1.1}} & \textbf{68.50}  \textls[-140]{\scalebox{0.8}{$\pm$ 1.3}} & \textbf{60.97}  \textls[-140]{\scalebox{0.8}{$\pm$ 0.7}}\\
\hline
\multicolumn{5}{c}{}  \\[-3pt]
\multicolumn{5}{c}{ResNet18}  \\
 \hline
Method   & P \hspace{-3pt}$\rightarrow{}$\hspace{-4pt} A & P \hspace{-3pt}$\rightarrow{}$\hspace{-4pt} C & P \hspace{-3pt}$\rightarrow{}$\hspace{-4pt} S &  Avg. \\ \hline \hline
ERM & 64.10 & 23.60 & 29.10 & 38.90 \\
SagNet \cite{SagNet} &  \underline{69.80} & 35.10 & 40.70 & 48.50 \\
SelfReg \cite{SelfReg} &  67.72 & 28.97 & 33.71 & 43.46 \\
XDED \cite{lkk22} & \textbf{71.40} & \underline{54.30} & 51.50 & 59.10 \\
ITTA \cite{ITTA} & 66.50 & 52.20 & \underline{63.80} & 60.80 \\
MCL \cite{MCL}  &  - & - & - & 59.60 \\
ProRandC \cite{ProRandConv}   & - & - & - & \underline{62.89} \\
CADA \cite{CADA}  & - & - & - & 56.65 \\ 
ABA$_{5l}$ \cite{ABA} & - & - & - & 59.04 \\ \hline
\tiny \textls[-140]{\checkmark$\hat{I}S\righttarrow I^{.75}S$} (Ours) & 67.97 \textls[-140]{\scalebox{0.8}{$\pm$ 0.6}} & \textbf{54.45} \textls[-140]{\scalebox{0.8}{$\pm$ 1.2}} & \textbf{74.13} \textls[-140]{\scalebox{0.8}{$\pm$ 0.7}} & \textbf{65.85} \textls[-140]{\scalebox{0.8}{$\pm$ 0.5}} \\
\hline
\end{tabular}
\end{center}
\vspace{-10pt}
\caption{\textbf{PACS.}
\label{tab:PACS}}
\fontsize{6}{6}\selectfont
\def\arraystretch{1.15}
\begin{tabular}{l@{\xlsp}l@{\xlsp}l@{\xlsp} l@{\xtssp} }
\hline
Method &  Hospital $4$ & Hospital $5$ &  Avg. \\ \hline \hline
ERM &  90.58 & 82.26 & 86.42  \\
AdvBNN \cite{AdvBNN}  & 87.30 & 80.79 & 84.04 \\
AugMix \cite{AugMixpaper}  & 85.92 & 84.86 & 85.39 \\
AugMax \cite{AugMax} & 79.61 & 85.51 & 82.56 \\
RandConv \cite{RandConv} & 90.64 & 84.75 & 87.70 \\
ABA$_{5l+A}$ \cite{ABA}  & 91.85 & 87.92 & 89.88 \\
 \hline
\tiny \textls[-140]{\checkmark $IS\righttarrow I$} (Ours)  &  \underline{92.20} \textls[-140]{\scalebox{0.8}{$\pm$ 1.9}} &  \underline{94.91} \textls[-140]{\scalebox{0.8}{$\pm$ 0.9}} &  \underline{93.56} \textls[-140]{\scalebox{0.8}{$\pm$ 1.4}} \\
\tiny \textls[-140]{$\hat{I}S\righttarrow I^{.75}S$} (Ours)  &  \textbf{92.88} \textls[-140]{\scalebox{0.8}{$\pm$ 0.7}} &  \textbf{95.86} \textls[-140]{\scalebox{0.8}{$\pm$ 0.2}} &  \textbf{94.37} \textls[-140]{\scalebox{0.8}{$\pm$ 0.3}} \\
\hline
\end{tabular}

\caption{\textbf{Camelyon17 with ResNet50.}
\label{tab:camelyon}}
\end{subtable}
\end{minipage}
\vspace{-10pt}
\centering
\caption{Comparison with the state-of-the-art on Digits (a), Mini-DomainNet (b), PACS (c), and Camelyon17 (d). The source domains are MNIST, real, photo, and hospitals 1-3, respectively. Each column represents a different target domain. Methods evaluated by us are denoted with a $\dagger$, and the variant of our method chosen by $V_A$ is denoted by \checkmark. ERM corresponds to $I\righttarrow I$.
\label{tab:soa}
\vspace{-15pt}
}
\end{table}

\paragraph{Comparison with the state-of-the-art} is presented in Table~\ref{tab:soa}. Whenever available, we include the reported results from the relevant publications. For the experiments on Mini-DomainNet we use the official code of each method.
The results reported for our method were obtained by fully automated
tuning of the learning rate and method selection among the $\hat{I}S\righttarrow I^wS$ variants on the augmented validation set. \emph{To our knowledge, no previously reported performance has resulted from such a validation process.}
State-of-the-art results are achieved on all four datasets.

\section{Conclusions}
\label{sec:conclusion}
We show that independent augmentations of the validation set allow for better model selection and hyper-parameter tuning in single-source domain generalization.
The proposed augmented validation enables the prediction of the test performance for prior methods and the proposed family of methods. Compared to the standard validation practice on the training distribution, the proposed validation method results in significant performance gains in real-world method selection over six domain generalization approaches. We expect this contribution to be valuable for future comparisons and to help researchers avoid the malpractice of tuning on the test set.
We further demonstrate that shape extraction in the form of cleaned edge maps is a solid tool for enforcing shape bias and enhancing the domain generalization ability of deep classifiers. State-of-the-art performance is achieved on several benchmarks by a method selected and with hyper-parameters tuned in a fully automated manner on the augmented validation set.

\textbf{Acknowledgments:} This work was supported by the Junior Star GACR GM 21-28830M, the Czech Technical University in Prague grant No. SGS23/173/OHK3/3T/13, and the CTU institutional support (Future fund). We also thank Nikolaos-Antonios Ypsilantis for the fruitful conversations and insightful discussions.
\clearpage

{\fontsize{9}{10}\selectfont
\bibliographystyle{ieee_fullname}
\bibliography{bib}
}
\appendix
\section{Appendix}

In this appendix, we provide additional information about the datasets used and implementation details both for the proposed and the literature methods utilized in our experiments. Additionally, we present qualitative examples of the data augmentations and the shape extraction processes used in this work. Finally, we include additional experimental results that could not fit in the main paper, along with tables containing the numerical data of our figures.

\subsection{Dataset Details}
\paragraph{Digits dataset details.} It is a collection of five digit-recognition datasets: MNIST, MNIST-M, SVHN, SYN, and USPS. MNIST-M combines the original MNIST handwritten digit database with random patches of the BSDS500 dataset. SVHN is a dataset of real-world house number images obtained by Google Street View. SYN is a synthetic dataset created from different Windows fonts after applying geometric transformations and blurring. USPS is a dataset of scanned digits from U.S. Postal Service envelopes. To compare with the literature, we use only the first $10,000$ images of the MNIST training set. We choose to use the $90\%$ for training and the $10\%$ for validation. We are evaluating on the test set of the rest domains.

\paragraph{PACS dataset details.} It is a domain generalization dataset that includes four domains: photo, art paintings, cartoon, and sketch. It consists of seven classes and $9,991$ images. In the experiments presented in the main paper, the photo domain is used as the source, and the remaining domains are used for evaluation. This appendix provides an additional experiment where each domain is used as the source. The photo domain consists of $1,670$ images. We use the official partition of $10\%$ for validation and the rest $90\%$ for training.

\paragraph{Mini-DomainNet dataset details.} It is a subset of the domain generalization dataset DomainNet~\cite{pbx19}. It consists of $140,006$ images, $126$ classes, and four domains: clipart, painting, real, and sketch. We use the real domain as the source, and we evaluate on the rest. The real domain has $64,979$ images and we are using the official split that includes $58,482$ images for training and $6,497$ images for validation.

\begin{figure}
\centering
\resizebox{1\columnwidth}{!}{%
\setlength{\fboxsep}{0pt}%
\setlength{\fboxrule}{0.5pt}%
\scriptsize
\begin{tabular}{c@{}c@{}c@{}c@{}c@{}c@{}c@{}}
& \multicolumn{3}{c}{Elephant} &  \multicolumn{3}{c}{House} \\[0.2cm] 
& Art Painting &  Cartoon  &  Sketch & Art Painting &  Cartoon  &  Sketch  \\[0.2cm] 
\rotatebox{90}{Original}\hspace{0.1cm} &
\fcolorbox{black}{black}{\includegraphics[height=1.4cm]{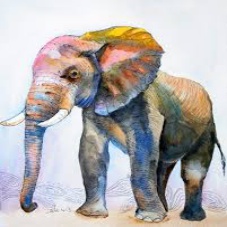}}\hspace{0.1cm} &
\fcolorbox{black}{black}{\includegraphics[height=1.4cm]{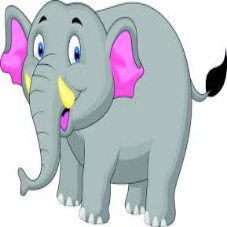}}\hspace{0.1cm} &
\fcolorbox{black}{black}{\scalebox{-1}[1]{\includegraphics[height=1.4cm]{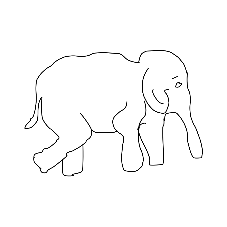}}}\hspace{0.1cm} &
\fcolorbox{black}{black}{\includegraphics[height=1.4cm]{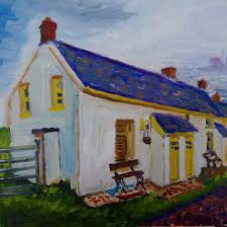}}\hspace{0.1cm} &
\fcolorbox{black}{black}{\includegraphics[height=1.4cm]{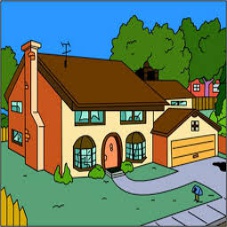}}\hspace{0.1cm} &
\fcolorbox{black}{black}{\includegraphics[height=1.4cm]{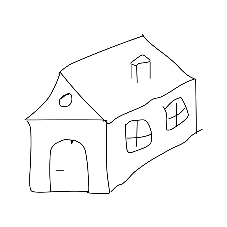}}\hspace{0.1cm}
\\[-0.02cm] 
\rotatebox{90}{Sobel}\hspace{0.1cm} &
\fcolorbox{black}{black}{\includegraphics[height=1.4cm]{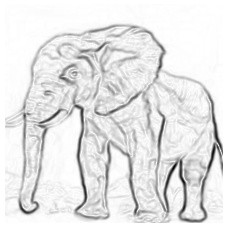}}\hspace{0.1cm} &
\fcolorbox{black}{black}{\includegraphics[height=1.4cm]{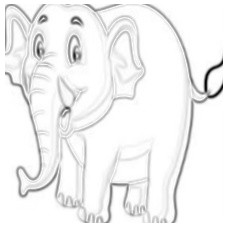}}\hspace{0.1cm} &
\fcolorbox{black}{black}{\scalebox{-1}[1]{\includegraphics[height=1.4cm]{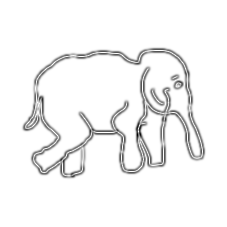}}}\hspace{0.1cm} &
\fcolorbox{black}{black}{\includegraphics[height=1.4cm]{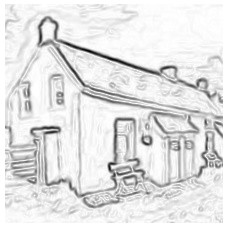}}\hspace{0.1cm} &
\fcolorbox{black}{black}{\includegraphics[height=1.4cm]{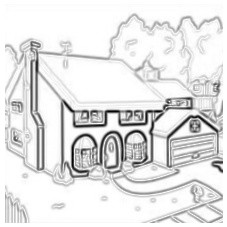}}\hspace{0.1cm} &
\fcolorbox{black}{black}{\includegraphics[height=1.4cm]{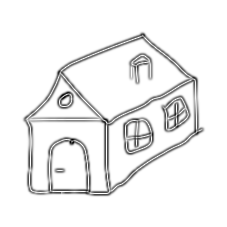}}\hspace{0.1cm}
\\[-0.02cm] 
\rotatebox{90}{BTE}\hspace{0.1cm} &
\fcolorbox{black}{black}{\includegraphics[height=1.4cm]{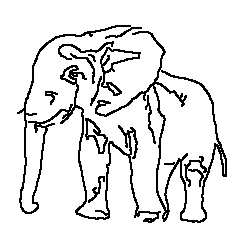}}\hspace{0.1cm} &
\fcolorbox{black}{black}{\includegraphics[height=1.4cm]{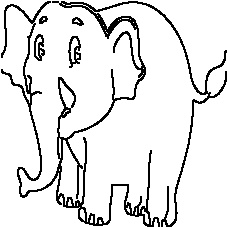}}\hspace{0.1cm} &
\fcolorbox{black}{black}{\scalebox{-1}[1]{\includegraphics[height=1.4cm]{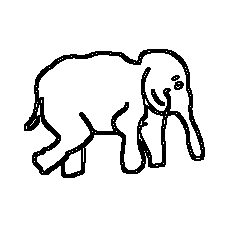}}}\hspace{0.1cm} &
\fcolorbox{black}{black}{\includegraphics[height=1.4cm]{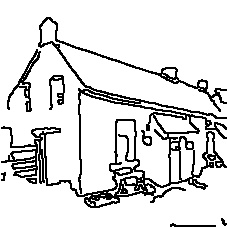}}\hspace{0.1cm} &
\fcolorbox{black}{black}{\includegraphics[height=1.4cm]{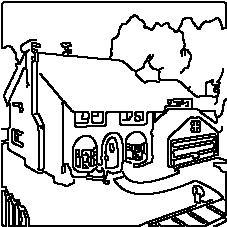}}\hspace{0.1cm} &
\fcolorbox{black}{black}{\includegraphics[height=1.4cm]{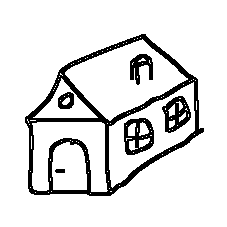}}\hspace{0.1cm}
\\[-0.02cm]

\end{tabular}

}

\caption{The output of the shape extraction process for BTE and Sobel-based edge maps from the different target domains in the PACS dataset. BTE removes texture cues more effectively, making it a better choice for increasing the shape bias. \label{fig:shape_target}}
\vspace{-5pt}
\end{figure}

\begin{figure}
\centering
\resizebox{1\columnwidth}{!}{%
\setlength{\fboxsep}{0pt}%
\setlength{\fboxrule}{0.5pt}%
\scriptsize
\begin{tabular}{c@{}c@{}c@{}c@{}c@{}c@{}c@{}}
& Dog & Elephant & Giraffe & Guitar & Horse & House \\[0.2cm] 
\rotatebox{90}{Original}\hspace{0.1cm} &
\fcolorbox{black}{black}{\includegraphics[height=1.4cm]{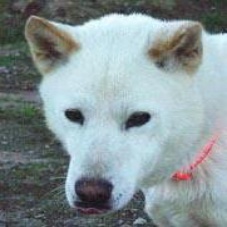}}\hspace{0.1cm} &
\fcolorbox{black}{black}{\includegraphics[height=1.4cm]{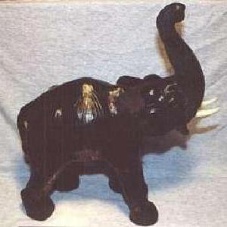}}\hspace{0.1cm} &
\fcolorbox{black}{black}{\includegraphics[height=1.4cm]{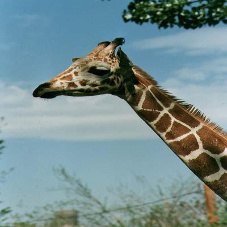}}\hspace{0.1cm} &
\fcolorbox{black}{black}{\includegraphics[height=1.4cm]{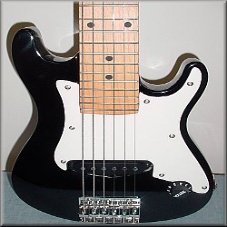}}\hspace{0.1cm} &
\fcolorbox{black}{black}{\includegraphics[height=1.4cm]{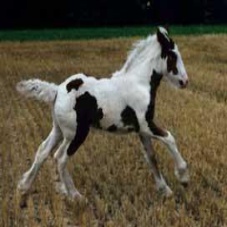}}\hspace{0.1cm} &
\fcolorbox{black}{black}{\includegraphics[height=1.4cm]{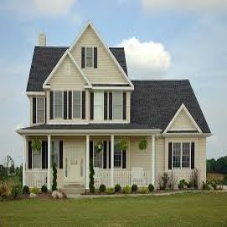}}\hspace{0.1cm} \\

\rotatebox{90}{random BTE 1}\hspace{0.1cm} &
\fcolorbox{black}{black}{\includegraphics[height=1.4cm]{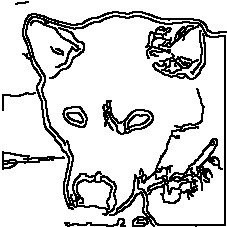}}\hspace{0.1cm} &
\fcolorbox{black}{black}{\includegraphics[height=1.4cm]{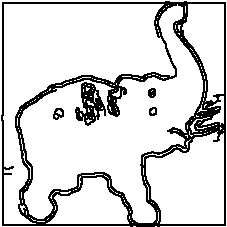}}\hspace{0.1cm} &
\fcolorbox{black}{black}{\includegraphics[height=1.4cm]{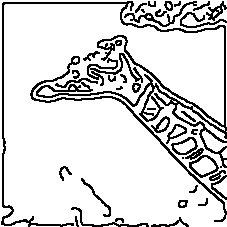}}\hspace{0.1cm} &
\fcolorbox{black}{black}{\includegraphics[height=1.4cm]{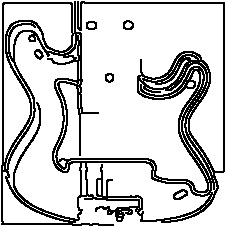}}\hspace{0.1cm} &
\fcolorbox{black}{black}{\includegraphics[height=1.4cm]{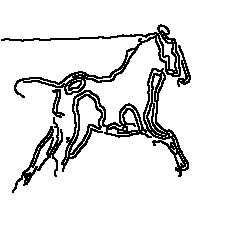}}\hspace{0.1cm} &
\fcolorbox{black}{black}{\includegraphics[height=1.4cm]{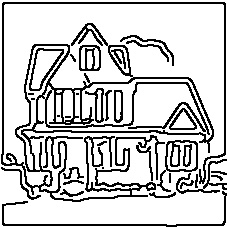}}\hspace{0.1cm} \\
\rotatebox{90}{random BTE 2}\hspace{0.1cm} &
\fcolorbox{black}{black}{\includegraphics[height=1.4cm]{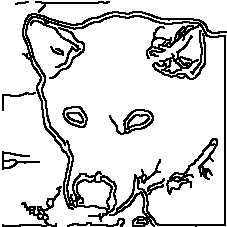}}\hspace{0.1cm} &
\fcolorbox{black}{black}{\includegraphics[height=1.4cm]{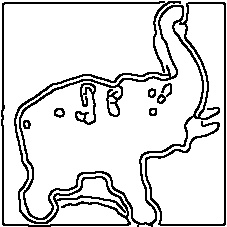}}\hspace{0.1cm} &
\fcolorbox{black}{black}{\includegraphics[height=1.4cm]{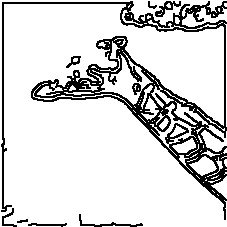}}\hspace{0.1cm} &
\fcolorbox{black}{black}{\includegraphics[height=1.4cm]{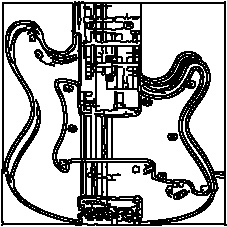}}\hspace{0.1cm} &
\fcolorbox{black}{black}{\includegraphics[height=1.4cm]{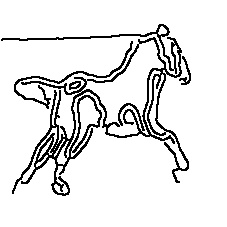}}\hspace{0.1cm} &
\fcolorbox{black}{black}{\includegraphics[height=1.4cm]{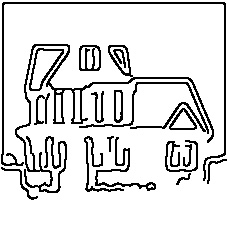}}\hspace{0.1cm} \\
\rotatebox{90}{random BTE 3}\hspace{0.1cm} &
\fcolorbox{black}{black}{\includegraphics[height=1.4cm]{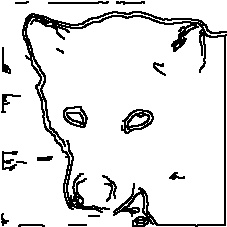}}\hspace{0.1cm} &
\fcolorbox{black}{black}{\includegraphics[height=1.4cm]{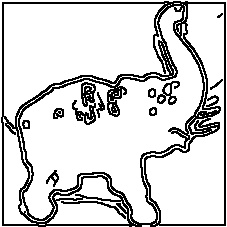}}\hspace{0.1cm} &
\fcolorbox{black}{black}{\includegraphics[height=1.4cm]{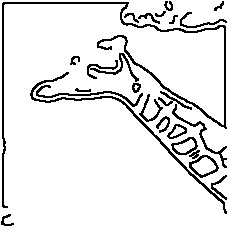}}\hspace{0.1cm} &
\fcolorbox{black}{black}{\includegraphics[height=1.4cm]{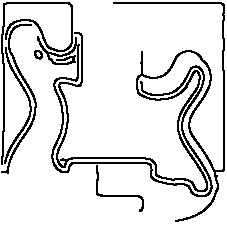}}\hspace{0.1cm} &
\fcolorbox{black}{black}{\includegraphics[height=1.4cm]{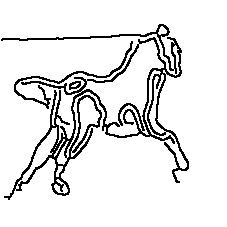}}\hspace{0.1cm} &
\fcolorbox{black}{black}{\includegraphics[height=1.4cm]{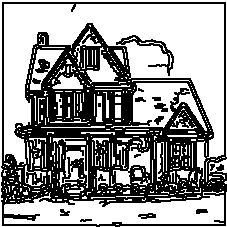}}\hspace{0.1cm} \\
\\[-0.02cm]

\end{tabular}

}

\vspace{-5pt}
\caption{The output of the shape extraction process for BTE edge maps on images of the photo domain in PACS during training. BTE introduces randomization at multiple steps of edge detection, enriching the training set with edge maps that vary in level of detail. \label{fig:shape_source_2}}
\vspace{-25pt}
\end{figure}

\paragraph{NICO++ dataset details.} It is an out-of-distribution generalization dataset comprising natural images, where the following contexts serve as the domains: autumn, dim light, grass, outdoor, rock, and water. NICO++ consists of $60$ classes and $88,866$ images.

\paragraph{Camelyon17 dataset details.} It is a medical dataset focused on tumor detection, with data from five different hospitals. Data from hospitals $1$, $2$, and $3$ are treated as the source domain, and data from hospitals $4$ and $5$ are the target. Camelyon17 consists of two classes: cancerous and non-cancerous tissue, and it contains $455,954$ images.

\paragraph{$\boldsymbol{16}$-class-ImageNet dataset details.} The $16$-class-ImageNet dataset \cite{gtr+18} is a subset of the ImageNet dataset that maps $231$ of the original classes to $16$ new ones, closer to the level of abstraction a human could guess. Although the $16$-class-ImageNet consists of $213,555$ images, we sub-sample to $500$ images per class. We test the trained models to the texture-shape cue conflict stimuli dataset \cite{grm+19}, which is a dataset that shares the same $16$ classes and consists of $1,280$ images. 

\subsection{Implementation Details}

\paragraph{Shape extraction.} The pipeline for BTEs and their randomization is adopted from~\cite{etc22} with the exception that we use Sobel instead of the learnable edge detectors, eliminating the need for additional training data. The pipeline is as follows: First, the image is blurred using a Gaussian filter with kernel size $5$ and sigma equal to $1.0$. Next, the Sobel operator is applied for edge detection, followed by non-maximum suppression to thin the edge map. Finally, the edge map is binarized using adaptive hysteresis. The upper and lower bounds of hysteresis are chosen as $1.5t$ and $0.5t$, respectively, where the threshold $t$ is selected using Otsu's method on the edge map before thinning.  

In the randomized variant used for training, the standard deviation of the Gaussian blurring is chosen randomly from $0$,$1$, and $2$, with $0$ corresponding to no blur. The thresholding method is randomly picked among Yen~\cite{ycc95}, Otsu~\cite{o79}, Isodata~\cite{rc78}, Li~\cite{ll93}, and the mean method~\cite{g93}. Additional random noise is introduced in both the threshold value $t$ and the hysteresis bounds, enriching the training set.

In the variant using Sobel edge maps of Table~\ref{tab:training_ablations}, the process is simplified in blurring and applying the Sobel operator. Randomization is only through the standard deviation of the Gaussian blurring. Examples of Sobel-based edge maps and BTEs are shown in Figures \ref{fig:shape_target} and \ref{fig:shape_source_2}.

\paragraph{Implementation details for our approach.} The \emph{basic augmentations} include cropping with relative size in $[0.8,1.0]$, an aspect ratio in $[\frac{3}{4},\frac{4}{3}]$, resizing to $224\times224$, and horizontal flipping with a probability of $0.5$. Digits is an exception where the resize is $32 \times 32$, and the flipping is skipped as it conflicts with the task. 
The \emph{extra augmentations} from the ImgAug library are from the following groups: arithmetic, artistic, blur, color, contrast, convolutional, edges, geometric, segmentation, and weather.
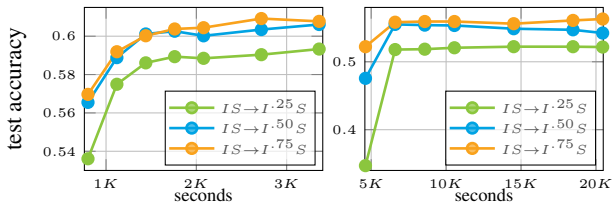
\begin{figure}[t]
\centering
\begin{tabular}{cc}
\hspace{-12.5pt}
\definecolor{myblue}{RGB}{72, 133, 237}
\pgfplotsset{every tick label/.append style={font=\tiny}}
\begin{tikzpicture}[define rgb/.code={\definecolor{mycolor}{RGB}{#1}}, rgb color/.style={define rgb={#1},mycolor}]
\small
\begin{axis}[
    title style = {yshift=-5pt},
    width=0.57\linewidth,
    height=0.45\linewidth,
    ymin=0.53,
    ymax=0.615,
    xmin=760,xmax=3400,
    xlabel style = {font=\scriptsize, yshift = 6pt},
    ylabel = {test accuracy},
    xlabel = {seconds},
    xtick={1000, 2000, 3000},
    xticklabels={$1K$, $2K$, $3K$}, 
    scaled x ticks = true, 
    grid=both,
    legend pos=south east,
    legend style={cells={anchor=west}, font =\tiny, fill opacity=0.7, row sep=-2.9pt, inner sep=1pt},
    ]

\addplot[mark=*,line width=1pt,LimeGreen, error bars/.cd, 
    y fixed,
    y dir=both, 
    y explicit] table[x=Sec,y=test_25_75
]{./data/appendix/method_percent_vs_time_pacs.csv};
\addlegendentry{$IS\righttarrow I^{.25}S$};

\addplot[mark=*,line width=1pt,Cerulean, error bars/.cd, 
    y fixed,
    y dir=both, 
    y explicit] table[x=Sec,y=test_50_50
]{./data/appendix/method_percent_vs_time_pacs.csv};
\addlegendentry{$IS\righttarrow I^{.50}S$};

\addplot[mark=*,line width=1pt,YellowOrange, error bars/.cd, 
    y fixed,
    y dir=both, 
    y explicit] table[x=Sec,y=test_75_25
]{./data/appendix/method_percent_vs_time_pacs.csv};
\addlegendentry{$IS\righttarrow I^{.75}S$};

\end{axis}
\end{tikzpicture}

& \hspace{-17pt}
\pgfplotsset{every tick label/.append style={font=\tiny}}
\begin{tikzpicture}[define rgb/.code={\definecolor{mycolor}{RGB}{#1}}, rgb color/.style={define rgb={#1},mycolor}]
\small
\begin{axis}[
    title style = {yshift=-5pt},
    width=0.57\linewidth,
    height=0.45\linewidth,
    ymin=0.34,
    ymax=0.58,
    xmin=4580,
    xmax=20485,
    xlabel style = {font=\scriptsize, yshift = 6pt},
    xlabel = {seconds},
    xtick={5000, 10000, 15000, 20000},
    xticklabels={$5K$, $10K$, $15K$, $20K$}, 
    scaled x ticks = true, 
    grid=both,
    legend pos=south east,
    legend style={cells={anchor=west}, font =\tiny, fill opacity=0.7, row sep=-2.9pt, inner sep=1pt},
    scaled x ticks = false
    ]

\addplot[mark=*,line width=1pt,LimeGreen, error bars/.cd, 
    y fixed,
    y dir=both, 
    y explicit] table[x=Sec,y=test_25_75
]{./data/appendix/method_percent_vs_time_minidn.csv};
\addlegendentry{$IS\righttarrow I^{.25}S$};

\addplot[mark=*,line width=1pt,Cerulean, error bars/.cd, 
    y fixed,
    y dir=both, 
    y explicit] table[x=Sec,y=test_50_50
]{./data/appendix/method_percent_vs_time_minidn.csv};
\addlegendentry{$IS\righttarrow I^{.50}S$};

\addplot[mark=*,line width=1pt,YellowOrange, error bars/.cd, 
    y fixed,
    y dir=both, 
    y explicit] table[x=Sec,y=test_75_25
]{./data/appendix/method_percent_vs_time_minidn.csv};
\addlegendentry{$IS\righttarrow I^{.75}S$};

\end{axis}
\end{tikzpicture}
\end{tabular}
\vspace{-5pt}
\caption{\textbf{Accuracy vs Time:} Training on PACS (left) and on Mini-DomainNet (right) with different number of BTEs in the batch. The training batch includes $64$ RGB images plus $0$, $8$, $16$, $24$, $32$, $48$, and $64$ BTEs (dots).
\label{fig:acc_vs_time}
\vspace{-5pt}
}
\end{figure}

\begin{figure}[t]
\centering
\begin{tikzpicture}
\small
\begin{axis}[
    title = {Spearman's corr. \textbf{\red{-0.44}}},
    width=0.6\linewidth,
    height=0.45\linewidth,
    xmin=90.0,
    xmax=102.0,
    ymin=25.0,
    ymax=70.0,
    ylabel style = {yshift = -3pt},
    xtick={90,95,100},
    title style = {yshift=-5pt},
    ylabel = {test accuracy},
    xlabel = {$V_S$ accuracy},
    legend pos=south west,
    legend style={cells={anchor=east}, font =\tiny, fill opacity=0.8, row sep=-2.5pt},
    ]
\addplot[only marks, mark=*, opacity=0.7,mark size=1.5,color=Tan, error bars/.cd, 
    y fixed,
    y dir=both, 
    y explicit] table[x=Source_Val,y=Source_Val_Rest
]{./data/appendix/small_v_a.csv};

\addplot[only marks, mark=*, opacity=0.7,mark size=1.5,color=Tan, error bars/.cd, 
    y fixed,
    y dir=both, 
    y explicit] table[x=Source_Val,y=Source_Val_Baseline
]{./data/appendix/small_v_a.csv};

\end{axis}
\end{tikzpicture}
\begin{tikzpicture}
\small
\begin{axis}[
    title = {Spearman's corr. \textbf{\green{0.80}}},
    width=0.6\linewidth,
    height=0.45\linewidth,
    xmin=62.0,
    xmax=85.0,
    ymin=25.0,
    ymax=70.0,
    xtick={60,65,...,85},
    title style = {yshift = -5pt},
    yticklabels=\empty,
    xlabel = {$V_a$ accuracy},
    legend pos=south west,
    legend style={cells={anchor=east}, font =\footnotesize, fill opacity=0.8, row sep=-2.5pt},
    ]
\addplot[only marks, mark=*, opacity=0.6,mark size=1.5,color=RoyalBlue, error bars/.cd, 
    y fixed,
    y dir=both, 
    y explicit] table[x=10_aug_ImgAug_Val,y=10_aug_Rest
]{./data/appendix/small_v_a.csv};

\addplot[only marks, mark=*, opacity=0.6,mark size=1.5,color=RoyalBlue, error bars/.cd, 
    y fixed,
    y dir=both, 
    y explicit] table[x=10_aug_ImgAug_Val,y=10_aug_Baseline
]{./data/appendix/small_v_a.csv};
\end{axis}
\end{tikzpicture}
\vspace{-5pt}
\caption{\textbf{Augmented validation without increasing the validation set size:} Correlation between the validation and test accuracy using the standard validation and the augmented one of the same size. This experiment shows that it is not the larger size of $V_A$ that makes the difference. The experiment is on the PACS dataset with a ResNet-18 as a backbone.\label{fig:abla_Va}}
\vspace{-15pt}
\end{figure}
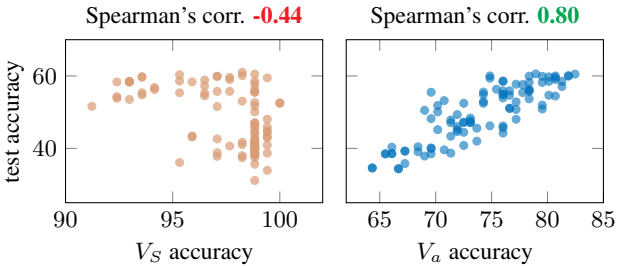
For our PACS, Digits, and Mini-DomainNet experiments, the \emph{learning rate} is tuned using a grid search among $33$ equidistant values on a logarithmic scale in the range of $[10^{-5}, 1]$. The \emph{loss weight} $\lambda$ is tuned using a grid search among 17 equidistant values in $[0, 1]$. The \emph{exponent} \emph{$w$} is a test-time parameter, and it is tuned among the values $0$, $0.25$, $0.5$, $0.75$, and $1.0$ for all of our experiments. The first and last values correspond to the variants $S$ and $I$, respectively. 
Experiments on PACS and Digits for the comparison with the state-of-the-art are repeated $30$ times.
In all other experiments on PACS, Digits, and Mini-DomainNet, we use $5$, $5$, and $3$ seeds, respectively.

Camelyon17 and NICO++ require longer training because of the larger size of the former and the randomly initialized training of the latter. Therefore, we perform a learning rate grid search for $9$ equidistant values on a logarithmic scale, in the range of $[10^{-5}, 1]$, while we train for $3$ different seeds.

For the $16$-class-ImageNet experiment, we perform a grid search for our $IS$ method's loss weight $\lambda$ for $16$ different values in the range of $[0.0625, 1]$.

We always use stochastic gradient descent with an exponential scheduler that decreases the learning rate by two magnitudes by the end of the training. We tune the number of epochs and the learning rate jointly for the $IS \righttarrow IS$ variant. This experiment  provides us with the number of epochs to use for all variants, which remains fixed for all the follow-up experiments described in the main paper. We train our models for $10$, $40$, $50$, $300$, $300$, and $700$ epochs on Camelyon17, Mini-DomainNet, $16$-class-ImageNet, Digits, PACS, and NICO++ respectively. 

\begin{table}[t]
\centering
\begin{center}
\resizebox{\columnwidth}{!}{%
\begin{tabular}{l@{\msp}l@{\msp}l@{\msp}l@{\msp}l@{\msp}l@{\msp}l@{\msp}l@{\msp}}
 \hline
Method   & Art & Cartoon & Photo & Sketch &  Avg. \\ \hline \hline
ERM      & 68.80 & 70.00 & 38.90 & 39.40 & 54.30 \\
JiGen \cite{JiGen} & 67.70 & 72.23 & 41.70 & 36.83 & 54.60 \\
ADA \cite{ADA} & 72.43 & 71.97 & 44.63 & 45.73 & 58.70 \\
SelfReg \cite{SelfReg} &  72.59 & 76.56 & 43.46 & 45.76 & 59.59 \\
SagNet \cite{SagNet} &  73.20 & 75.67 & 48.53 & 50.07 & 61.90 \\
GeoTexAug \cite{GeoTexAug} & 72.07 & 78.70 & 49.07 & 59.97 & 65.00 \\
L2D \cite{Wlq+21} & 76.91 & 77.88 & 52.29 & 53.66 & 65.18 \\
XDED \cite{lkk22} & 76.50 & 77.20 & 59.10 & 53.10 & 66.50 \\
CADA \cite{CADA}  & 76.33 & \underline{79.08} & 61.59 & 56.65 & 68.41 \\
ITTA \cite{ITTA} & 74.60 & 77.10 & 60.80 & \textbf{61.20} & 68.40 \\
ProRandC \cite{ProRandConv}   & 76.98 & 78.54 & \underline{62.89} & 57.11 & 68.88 \\
MCL \cite{MCL}  &  \underline{77.13} & \textbf{80.14} & 62.55 & \underline{59.60} & \underline{69.86} \\
 ABA$_{3l}$ & 75.34 & 77.49 & 58.86 & 53.76 & 66.36 \\
 \hline
$\hat{I}S\righttarrow I^{.75}S$ (Ours) & \textbf{80.67} \scalebox{0.8}{$\pm$ 0.4} & 76.53 \scalebox{0.8}{$\pm$ 1.1} & \textbf{65.85} \scalebox{0.8}{$\pm$ 0.5} & 58.41 \scalebox{0.8}{$\pm$ 1.3} & \textbf{70.37} \scalebox{0.8}{$\pm$ 0.5} \\
\hline
\end{tabular}
}
\end{center}

\vspace{-5pt}
\caption{\textbf{Comparison with state-of-the-art approaches on PACS with a ResNet18 backbone.} Each column corresponds to a different source domain, reporting average performance when testing on the three remaining domains as target domains. 
\label{tab:PACS_all_domains}
\vspace{-15pt}
}
\end{table}

\begin{table*}[t]
\centering
\fontsize{8}{9}\selectfont
\def\arraystretch{1.1}
\begin{tabular}{l@{\ssp}l@{\ssp}r@{\lsp}l@{\ssp}r@{\lsp}l@{\ssp}r@{\lsp}l@{\ssp}r@{\lsp}l@{\ssp}r@{\lsp}l@{\ssp}r@{\lsp}l@{\ssp}r}
\hline
 &
  \multicolumn{2}{c}{PACS-ViT-S} &
  \multicolumn{2}{c}{PACS-RN18} &
  \multicolumn{2}{c}{MiniDN-RN18} &
  \multicolumn{2}{c}{MiniDN-Alexnet} &
  \multicolumn{2}{c}{NICO++-RN18} &
  \multicolumn{2}{c}{Digits-LeNet}&
  \multicolumn{2}{c}{Cam17-RN50} \\ \hline
  Val &
  Method &
  Acc &
  Method &
  Acc &
  Method &
  Acc &
  Method &
  Acc &
  Method &
  Acc &
  Method &
  Acc &
  Method &
  Acc \\ \hline \hline
$V_O$ &
  \tiny \textls[-140]{$\hat{I}S \righttarrow IS$} &
  75.68 &
  \tiny  \textls[-140]{$\hat{I}S \righttarrow I^{.75}S$} &
  66.19 &
  \tiny \textls[-140]{$\hat{I}S \righttarrow I^{.75}S$} &
  57.89 &
  \tiny  \textls[-140]{$\hat{I}S \righttarrow I^{.75}S$} &
  49.10 &
  \tiny   \textls[-140]{$\hat{I}S \righttarrow I^{.75}S$} &
  29.12 &
  \tiny  \textls[-140]{$\hat{I}S \righttarrow I^{.50}S$} &
  83.83 &
  \tiny  \textls[-140]{$\hat{I}S \righttarrow I$} &
  94.47 \\
$V_S$ &
  \tiny  \textls[-140]{$ties^*$} &
  68.70 &
  \tiny  \textls[-140]{$IS_{\times 2} \righttarrow I^{.50}S$} &
  51.75 &
  \tiny  \textls[-140]{$IS_{\times 2} \righttarrow I^{.75}S$} &
  52.98 &
 \tiny   \textls[-140]{$I \righttarrow I$} &
  39.82 &
  \tiny   \textls[-140]{$IS \righttarrow I^{.75}S$} &
  26.86 &
  \tiny  \textls[-140]{$IS_{sob} \righttarrow I$} &
  72.51&
  \tiny  \textls[-140]{$I \righttarrow I$} &
  78.73  \\
$V_A$ &
  \tiny  \textls[-140]{$\hat{I}S \righttarrow I^{.75}S$} &
  74.48 &
 \tiny   \textls[-140]{$\hat{I}S \righttarrow I^{.75}S$} &
  65.85 &
 \tiny   \textls[-140]{$\hat{I}S \righttarrow I^{.75}S$} &
  57.35 &
 \tiny   \textls[-140]{$\hat{I}S \righttarrow I^{.75}S$} &
  48.85 &
  \tiny   \textls[-140]{$\hat{I}S \righttarrow I^{.75}S$} &
  29.12 &
 \tiny   \textls[-140]{$\hat{I}S \righttarrow I^{.75}S$} &
  82.61 &
 \tiny   \textls[-140]{$IS \righttarrow I$} &
  93.56 \\ \hline
gain & & 5.78 & &  14.10& &   4.37 & &   9.03 & &  2.26 & & 10.10 & & 14.83   \\ \hline
\end{tabular}
\caption{\textbf{Method selection based on the validation set:} Test accuracy is reported after tuning and selecting the best method among all proposed variants according to different validation sets -- \ie, oracle, standard, and augmented. The method chosen and the performance gain between the augmented and the standard validation set are reported. In the case of PACS with a ViT-S model,  $V_S$ ties across seven variations, so we report the average.
\label{tab:global}
}
\vspace{-5pt}
\end{table*}

\begin{table}[t]
\centering
\begin{center}
\fontsize{6}{8}\selectfont
\def\arraystretch{1}
\begin{tabular}{@{\xssp}r@{\ssp}r@{\msp}c@{\lsp}c@{\msp}c@{\msp}c@{\lsp}c@{\lsp}c@{\msp}c@{\msp}c@{\lsp}c}
\hline 
\multicolumn{1}{c}{} &
  \multicolumn{1}{c}{} &
  \multicolumn{5}{c}{tune learning rate} &
  \multicolumn{4}{c}{tune loss weight $\lambda$} \\\hline 

&&
  Train $\righttarrow$ Test & $V_S$ & $V_A$ & $V_O$ & Gain & $V_S$ & $V_A$ & $V_O$ & Gain \\ \hline \hline 

 &
   &
  $IS \righttarrow I^{.25}S$ &
  58.0 &
  57.6 &
  61.0 &
  -0.4 &
  37.9 &
  56.1 &
  57.7 &
  18.3 \\
 &
   &
  $IS \righttarrow I^{.50}S$ &
  58.0 &
  59.7 &
  61.3 &
  1.7 &
  35.8 &
  57.1 &
  59.4 &
  21.4 \\
\multirow{-3}{*}{\rotatebox[origin=c]{90}{PACS}} &
  \multirow{-3}{*}{\rotatebox[origin=c]{90}{RN18}} &
  $IS \righttarrow I^{.75}S$ &
  56.7 &
  59.2 &
  61.3 &
  2.5 &
  49.0 &
  57.5 &
  60.4 &
  8.5 \\ \hline 
 &
   &
  $IS \righttarrow I^{.25}S$ &
  62.6 &
  65.4 &
  65.9 &
  2.8 &
  63.1 &
  64.9 &
  65.2 &
  1.8 \\
 &
   &
  $IS \righttarrow I^{.50}S$ &
  67.4 &
  68.5 &
  69.4 &
  1.1 &
  68.0 &
  68.1 &
  68.9 &
  0.1 \\
  
\multirow{-3}{*}{\rotatebox[origin=c]{90}{PACS}} &

  \multirow{-3}{*}{\rotatebox[origin=c]{90}{ViT-S}} &
  
  $IS \righttarrow I^{.75}S$ &
  
  69.4 &
  71.2 &
  71.5 &
  1.8 &
  70.4 &
  70.6 &
  71.2 &
  0.2 \\ \hline
 &
   &
   
  $IS \righttarrow I^{.25}S$ &
  45.3 &
  45.5 &
  45.9 &
  0.2 &
  43.6 &
  44.9 &
  45.4 &
  1.3 \\

 &
   &
     
  $IS \righttarrow I^{.50}S$ &
  46.9 &
  47.5 &
  48.3 &
  0.6 &
  45.9 &
  47.5 &
  47.9 &
  1.6 \\
  
\multirow{-3}{*}{\rotatebox[origin=c]{90}{MiniDN}} &

  \multirow{-3}{*}{\rotatebox[origin=c]{90}{Alexnet}} &
  
  $IS \righttarrow I^{.75}S$ &
  47.8 &
  47.8 &
  48.1 &
  0.0 &
  46.1 &
  48.0 &
  48.6 &
 1.9 \\ \hline

 &
   &
  $IS \righttarrow I^{.25}S$ &
  51.8 &
  51.6 &
  51.9 &
  -0.2 &
  47.3 &
  50.8 &
  51.0 &
  3.5 \\
 &
   &
  $IS \righttarrow I^{.50}S$ &
  55.0 &
  55.0 &
  55.5 &
  0.1 &
  48.1 &
  53.9 &
  54.6 &
  5.8 \\
\multirow{-3}{*}{\rotatebox[origin=c]{90}{MiniDN}} &
  \multirow{-3}{*}{\rotatebox[origin=c]{90}{RN18}} &
  $IS \righttarrow I^{.75}S$ &
  55.5 &
  55.3 &
  56.0 &
  -0.2 &
  50.1 &
  54.7 &
  54.9 &
  4.6 \\ \hline 
 &
 &
  $IS \righttarrow I^{.25}S$ &
  78.0 &
  78.8 &
  78.9 &
  0.8 &
  75.4 &
  76.0 &
  76.8 &
  0.7 \\
 &
 &
  $IS \righttarrow I^{.50}S$ &
  77.8 &
  78.8 &
  79.1 &
  0.9 &
  76.1 &
  76.3 &
  77.2 &
  0.2 \\
\multirow{-3}{*}{\rotatebox[origin=c]{90}{Digits}} &
  \multirow{-3}{*}{\rotatebox[origin=c]{90}{LeNet}} &
  $IS \righttarrow I^{.75}S$ &
  76.2 &
  78.6 &
  78.9 &
  2.4 &
  76.4 &
  76.7 &
  77.6 &
  0.3 \\ \hline 

 &
 &
  $IS \righttarrow I^{.25}S$ &
  23.4 &
  23.2 &
  23.6 &
  -0.2 &
  23.6 &
  23.7 &
  23.8 &
  0.1 \\
 &
 &
  $IS \righttarrow I^{.50}S$ &
  25.8 &
  25.7 &
  26.2 &
  -0.1 &
  26.0 &
  26.5 &
  26.7 &
  0.5 \\
\multirow{-3}{*}{\rotatebox[origin=c]{90}{NICO++}} &
  \multirow{-3}{*}{\rotatebox[origin=c]{90}{RN18}} &
  $IS \righttarrow I^{.75}S$ &
  26.1 &
  26.0 &
  26.6 &
  -0.1 &
  26.2 &
  26.8 &
  26.9 &
  0.6 \\ \hline 

 &
 &
  $IS \righttarrow I^{.25}S$ &
  92.0 &
  92.3 &
  92.4 &
  0.3 &
  92.1 &
  92.2 &
  92.4 &
  0.1 \\
 &
 &
  $IS \righttarrow I^{.50}S$ &
  93.3 &
  93.4 &
  93.9 &
  0.1 &
  93.7 &
  93.7 &
  93.9 &
  0.0 \\
\multirow{-3}{*}{\rotatebox[origin=c]{90}{Cam17}} &
  \multirow{-3}{*}{\rotatebox[origin=c]{90}{RN50}} &
  $IS \righttarrow I^{.75}S$ &
  93.8 &
  94.3 &
  94.5 &
  0.5 &
  94.1 &
  94.1 &
  94.5 &
  0.0 \\ \hline 
\multicolumn{4}{l}{Avg gain} &
   &
   &
  0.7 &
   &
   &
   &
  3.4 \\ \hline 
\end{tabular}
\end{center}

\vspace{-5pt}
\caption{\textbf{Learning rate and loss weight tuning per validation:} Test accuracy for our method variants is reported after tuning according to different validation sets -- \ie, oracle, standard, and augmented. The performance gain between the augmented versus the standard validation set is also presented. As expected, the method is more effective for tuning hyperparameters related to domain generalization, such as the shape loss weight.
\label{tab:lr_method_loss}
\vspace{-15pt}
}
\end{table}

\paragraph{Implementation details for the literature methods.} Regarding SelfReg, SagNet, L2D, and ACVC, we follow all the implementation details -- optimizer, schedulers, augmentations, and hyperparameters -- from the original works, except for the learning rate. To determine the number of training epochs, we first set the learning rate to the value reported in the publication of the respective method and tune the number of epochs to maximize validation accuracy. Ties are resolved by picking the smaller number, while we never go for more than 800 or 100 epochs on PACS and Mini-DomainNet, respectively. Once the number of epochs is tuned, we tune the learning rate to maximize validation accuracy. 

\subsection{Extra Experiments}

\paragraph{Performance vs Time:}
For the proposed validation $V_A$, there is no matter of time-performance trade-off. We argue that the standard validation $V_S$ is completely incapable of predicting test performance in the context of domain generalization. This can be seen from Figure~\ref{fig:sota_scatter}: PACS shows an accuracy drop of $22.2$ if $V_S$ is used over $V_A$. Methods that do not use the proposed augmentations in training, such as L2D and SelfReg, do not require the 2-fold cross-validation. In such cases, the time overhead of $V_A$ over $V_S$ is only the performance of a random augmentation. For methods that use augmentations, the extra time compared to $V_S$ is approximately doubled because of the 2-fold cross-validation.

For the proposed recognition method, the performance vs training time trade-off is shown in Figure \ref{fig:acc_vs_time}.
Even with roughly half of the training time, when only $25\%$ of the batch images have their BTEs (16 BTEs) used,  the test accuracy decrease is approximately $1\%$. The time measurements were conducted on a single Tesla A100 40GB GPU.

\paragraph{$V_A$ vs $V_S$: Effective because it is larger?}
The proposed validation method $V_A$ increases the variability in the validation set as well as the size of the validation set by a factor of 10, which is given by the 10 groups of augmentations. To demonstrate that the benefit does not come from the larger validation set, we create an additional set by augmenting each image only once by randomly picking one of the 10 augmentation groups per image. The result is an augmented set of the same size as the original validation set, which we denote as $V_a$. We perform the same experiments as in Figure~\ref{fig:all_datasets} for PACS with a ResNet-18, but we exclude all variations that use augmentations during training to avoid overestimation, as described in Figure~\ref{fig:abla_cros_val}.
Figure \ref{fig:abla_Va} shows that validation $V_a$ is still significantly better than $V_S$. 

\paragraph{Experiments with each domain as the source domain.} 
In Table \ref{tab:PACS_all_domains} we report the performance on the PACS dataset while using each domain as the source domain.
We consider this an invalid setup due to the ImageNet pre-training. The networks have seen both real images during the pre-training phase and also cartoons, artworks, or sketches during training, making it similar to an MSDG task. Additionally, evaluating on the photo domain no longer corresponds to testing on an unseen domain. Nevertheless, we report results following the example of the literature, and our method is, on average, the top performing. 
Training from scratch would make these setups valid for SSDG, but the literature lacks available results for comparison.

\paragraph{Method selection and hyperparameter tuning.}
We summarize the results of our experiments for method selection in Table \ref{tab:global} and for hyperparameter tuning in Table \ref{tab:lr_method_loss}. These tables contain the data presented in Figure~\ref{fig:radar} and Figure~\ref{fig:dot_plots}, respectively. While Figure~\ref{fig:dot_plots} includes only experiments with the variant $IS\righttarrow I^{.75}S$, Table \ref{tab:lr_method_loss} also contains the variants $IS\righttarrow I^{.25}S$ and $IS\righttarrow I^{.50}S$.

\begin{figure*}[t]
\centering
\vspace{-5pt}
\captionsetup{type=figure} 

\begin{minipage}{.5\textwidth}
\centering
\begin{subfigure}{\textwidth}
\resizebox{\columnwidth}{!}{%
\setlength{\fboxsep}{0pt}%
\setlength{\fboxrule}{0.5pt}%
\scriptsize
\begin{tabular}{c@{}c@{}c@{}c@{}c@{}c@{}c@{}}
& Dog & Elephant & Giraffe & Guitar & Horse & House  \\[0.2cm] 
\multicolumn{7}{c}{Original} \\
\rotatebox{90}{Original}\hspace{0.1cm} &
\fcolorbox{black}{black}{\includegraphics[width=1.4cm]{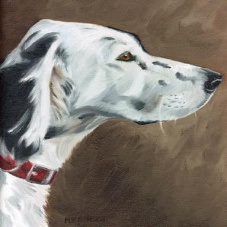}}\hspace{0.1cm} &
\fcolorbox{black}{black}{\includegraphics[width=1.4cm]{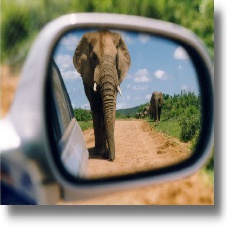}}\hspace{0.1cm} &
\fcolorbox{black}{black}{\includegraphics[width=1.4cm]{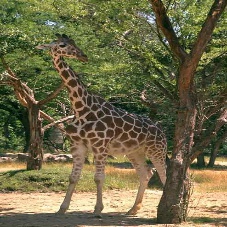}}\hspace{0.1cm} &
\fcolorbox{black}{black}{\includegraphics[width=1.4cm]{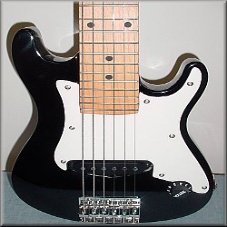}}\hspace{0.1cm} &
\fcolorbox{black}{black}{\includegraphics[width=1.4cm]{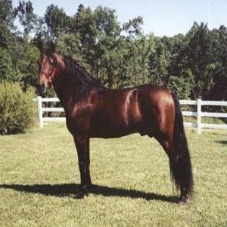}}\hspace{0.1cm} &
\fcolorbox{black}{black}{\includegraphics[width=1.4cm]{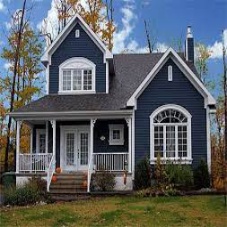}}\hspace{0.1cm} \\

\multicolumn{7}{c}{Arithmetic} \\
\rotatebox{90}{Add}\hspace{0.1cm} &
\fcolorbox{black}{black}{\includegraphics[width=1.4cm]{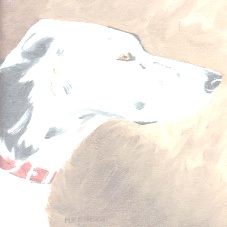}}\hspace{0.1cm} &
\fcolorbox{black}{black}{\includegraphics[width=1.4cm]{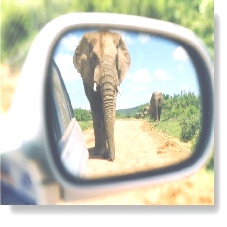}}\hspace{0.1cm} &
\fcolorbox{black}{black}{\includegraphics[width=1.4cm]{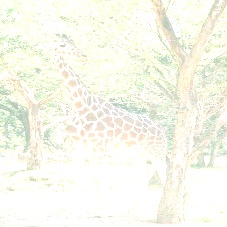}}\hspace{0.1cm} &
\fcolorbox{black}{black}{\includegraphics[width=1.4cm]{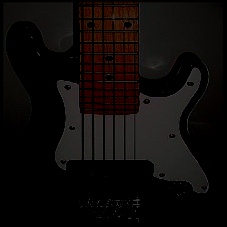}}\hspace{0.1cm} &
\fcolorbox{black}{black}{\includegraphics[width=1.4cm]{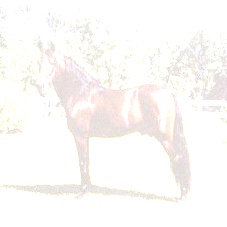}}\hspace{0.1cm} &
\fcolorbox{black}{black}{\includegraphics[width=1.4cm]{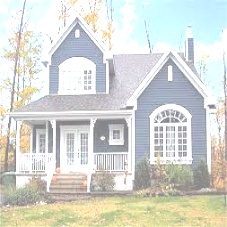}}\hspace{0.1cm} \\
\end{tabular}

}

\label{fig:aug}
\end{subfigure}
\end{minipage}%
\begin{minipage}{.5\textwidth}
\centering
\begin{subfigure}{\textwidth}
\resizebox{\columnwidth}{!}{%
\setlength{\fboxsep}{0pt}%
\setlength{\fboxrule}{0.5pt}%
\scriptsize
\begin{tabular}{c@{}c@{}c@{}c@{}c@{}c@{}c@{}}
& Dog & Elephant & Giraffe & Guitar & Horse & House  \\[0.2cm] 

\multicolumn{7}{c}{Arithmetic cont.} \\

\rotatebox{90}{Slt\&Ppr.}\hspace{0.1cm} &
\fcolorbox{black}{black}{\includegraphics[width=1.4cm]{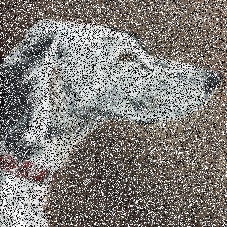}}\hspace{0.1cm} &
\fcolorbox{black}{black}{\includegraphics[width=1.4cm]{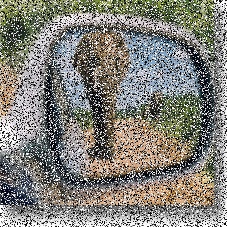}}\hspace{0.1cm} &
\fcolorbox{black}{black}{\includegraphics[width=1.4cm]{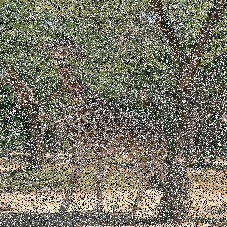}}\hspace{0.1cm} &
\fcolorbox{black}{black}{\includegraphics[width=1.4cm]{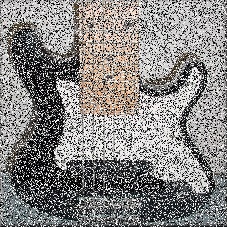}}\hspace{0.1cm} &
\fcolorbox{black}{black}{\includegraphics[width=1.4cm]{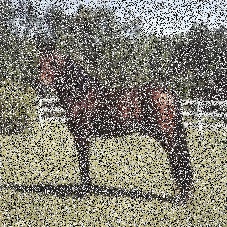}}\hspace{0.1cm} &
\fcolorbox{black}{black}{\includegraphics[width=1.4cm]{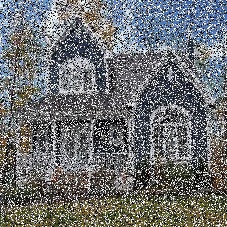}}\hspace{0.1cm} \\

\rotatebox{90}{CrsSlt\&Ppr.}\hspace{0.1cm} &
\fcolorbox{black}{black}{\includegraphics[width=1.4cm]{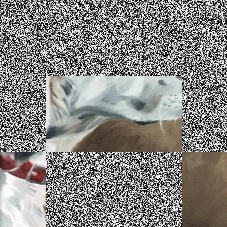}}\hspace{0.1cm} &
\fcolorbox{black}{black}{\includegraphics[width=1.4cm]{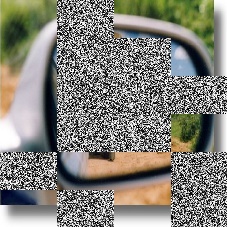}}\hspace{0.1cm} &
\fcolorbox{black}{black}{\includegraphics[width=1.4cm]{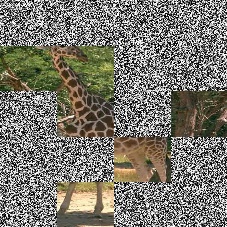}}\hspace{0.1cm} &
\fcolorbox{black}{black}{\includegraphics[width=1.4cm]{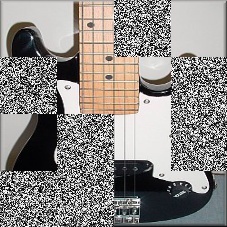}}\hspace{0.1cm} &
\fcolorbox{black}{black}{\includegraphics[width=1.4cm]{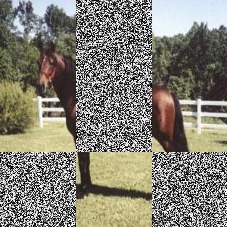}}\hspace{0.1cm} &
\fcolorbox{black}{black}{\includegraphics[width=1.4cm]{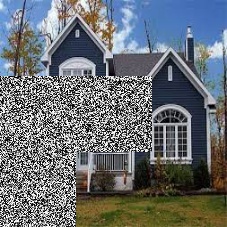}}\hspace{0.1cm} \\
\end{tabular}

}

\label{fig:aug2}
\end{subfigure}
\end{minipage}

\caption{Examples of augmentations used from each augmentation category.}
\label{fig:aug_combined1}
\vspace{-10pt}
\end{figure*}

\begin{figure*}[t]
\centering
\vspace{-5pt}
\captionsetup{type=figure} 

\begin{minipage}{.5\textwidth}
\centering
\begin{subfigure}{\textwidth}
\resizebox{\columnwidth}{!}{%
\setlength{\fboxsep}{0pt}%
\setlength{\fboxrule}{0.5pt}%
\scriptsize
\begin{tabular}{c@{}c@{}c@{}c@{}c@{}c@{}c@{}}
& Dog & Elephant & Giraffe & Guitar & Horse & House  \\[0.2cm] 

\multicolumn{7}{c}{Blur cont.} \\

\rotatebox{90}{Bilateral}\hspace{0.1cm} &
\fcolorbox{black}{black}{\includegraphics[width=1.4cm]{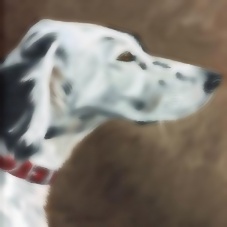}}\hspace{0.1cm} &
\fcolorbox{black}{black}{\includegraphics[width=1.4cm]{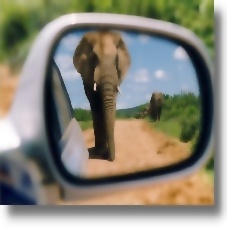}}\hspace{0.1cm} &
\fcolorbox{black}{black}{\includegraphics[width=1.4cm]{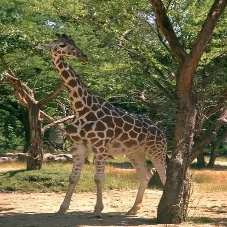}}\hspace{0.1cm} &
\fcolorbox{black}{black}{\includegraphics[width=1.4cm]{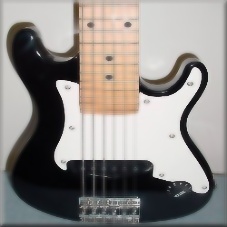}}\hspace{0.1cm} &
\fcolorbox{black}{black}{\includegraphics[width=1.4cm]{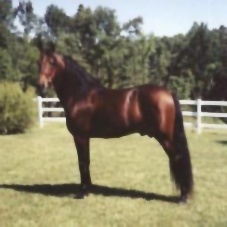}}\hspace{0.1cm} &
\fcolorbox{black}{black}{\includegraphics[width=1.4cm]{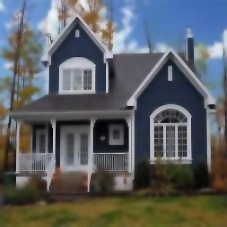}}\hspace{0.1cm} \\

\rotatebox{90}{Motion}\hspace{0.1cm} &
\fcolorbox{black}{black}{\includegraphics[width=1.4cm]{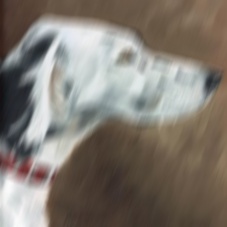}}\hspace{0.1cm} &
\fcolorbox{black}{black}{\includegraphics[width=1.4cm]{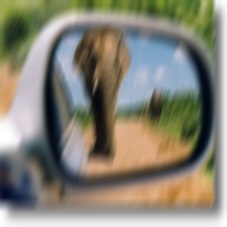}}\hspace{0.1cm} &
\fcolorbox{black}{black}{\includegraphics[width=1.4cm]{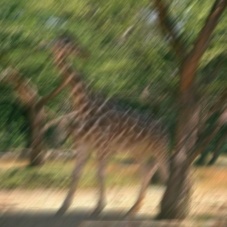}}\hspace{0.1cm} &
\fcolorbox{black}{black}{\includegraphics[width=1.4cm]{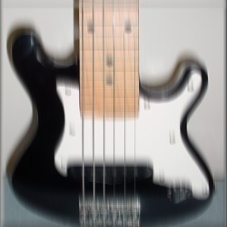}}\hspace{0.1cm} &
\fcolorbox{black}{black}{\includegraphics[width=1.4cm]{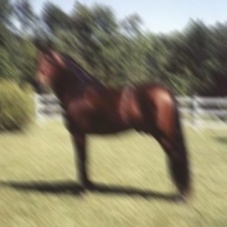}}\hspace{0.1cm} &
\fcolorbox{black}{black}{\includegraphics[width=1.4cm]{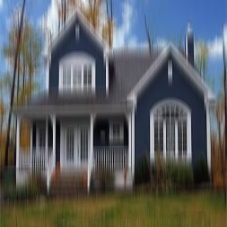}}\hspace{0.1cm} \\
\end{tabular}

}

\label{fig:aug3}
\end{subfigure}
\end{minipage}%
\begin{minipage}{.5\textwidth}
\centering
\begin{subfigure}{\textwidth}
\resizebox{\columnwidth}{!}{%
\setlength{\fboxsep}{0pt}%
\setlength{\fboxrule}{0.5pt}%
\scriptsize
\begin{tabular}{c@{}c@{}c@{}c@{}c@{}c@{}c@{}}
& Dog & Elephant & Giraffe & Guitar & Horse & House  \\[0.2cm] 

\multicolumn{7}{c}{Color Cont.} \\

\rotatebox{90}{AddHue\&Sat}\hspace{0.1cm} &
\fcolorbox{black}{black}{\includegraphics[width=1.4cm]{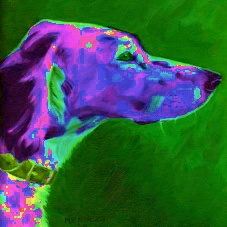}}\hspace{0.1cm} &
\fcolorbox{black}{black}{\includegraphics[width=1.4cm]{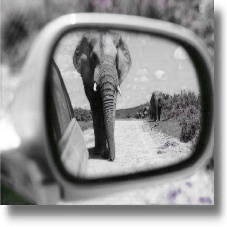}}\hspace{0.1cm} &
\fcolorbox{black}{black}{\includegraphics[width=1.4cm]{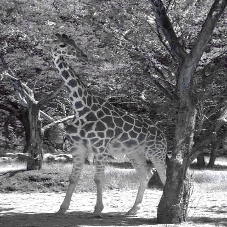}}\hspace{0.1cm} &
\fcolorbox{black}{black}{\includegraphics[width=1.4cm]{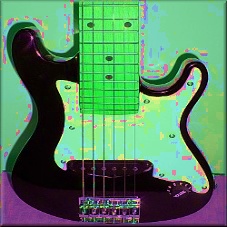}}\hspace{0.1cm} &
\fcolorbox{black}{black}{\includegraphics[width=1.4cm]{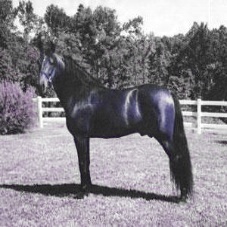}}\hspace{0.1cm} &
\fcolorbox{black}{black}{\includegraphics[width=1.4cm]{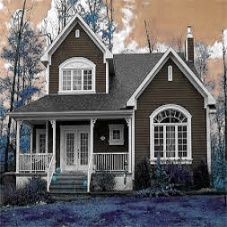}}\hspace{0.1cm} \\
\rotatebox{90}{AddHue}\hspace{0.1cm} &
\fcolorbox{black}{black}{\includegraphics[width=1.4cm]{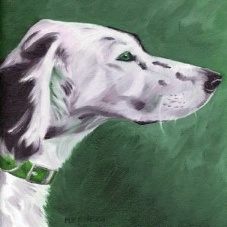}}\hspace{0.1cm} &
\fcolorbox{black}{black}{\includegraphics[width=1.4cm]{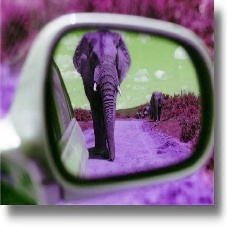}}\hspace{0.1cm} &
\fcolorbox{black}{black}{\includegraphics[width=1.4cm]{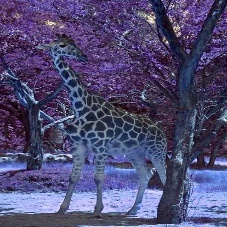}}\hspace{0.1cm} &
\fcolorbox{black}{black}{\includegraphics[width=1.4cm]{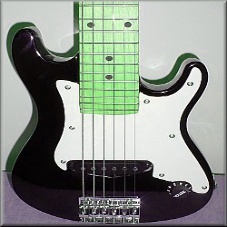}}\hspace{0.1cm} &
\fcolorbox{black}{black}{\includegraphics[width=1.4cm]{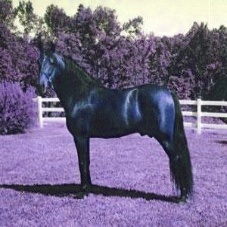}}\hspace{0.1cm} &
\fcolorbox{black}{black}{\includegraphics[width=1.4cm]{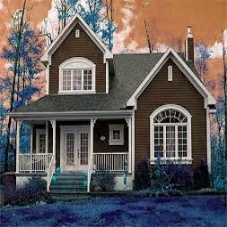}}\hspace{0.1cm} \\
\end{tabular}

}
\label{fig:aug4}
\end{subfigure}
\end{minipage}

\caption{Examples of augmentations used from each augmentation category.}
\label{fig:aug_combined2}
\vspace{-10pt}
\end{figure*}

\begin{figure*}[t]
\centering
\vspace{-5pt}
\captionsetup{type=figure} 

\begin{minipage}{.5\textwidth}
\centering
\begin{subfigure}{\textwidth}
\resizebox{\columnwidth}{!}{%
\setlength{\fboxsep}{0pt}%
\setlength{\fboxrule}{0.5pt}%
\scriptsize
\begin{tabular}{c@{}c@{}c@{}c@{}c@{}c@{}c@{}}
& Dog & Elephant & Giraffe & Guitar & Horse & House  \\[0.2cm] 

\multicolumn{7}{c}{Contrast cont.} \\

\rotatebox{90}{CLAHE}\hspace{0.1cm} &
\fcolorbox{black}{black}{\includegraphics[width=1.4cm]{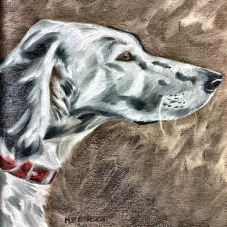}}\hspace{0.1cm} &
\fcolorbox{black}{black}{\includegraphics[width=1.4cm]{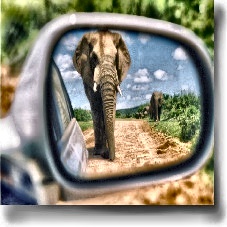}}\hspace{0.1cm} &
\fcolorbox{black}{black}{\includegraphics[width=1.4cm]{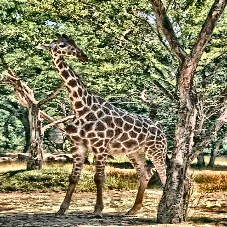}}\hspace{0.1cm} &
\fcolorbox{black}{black}{\includegraphics[width=1.4cm]{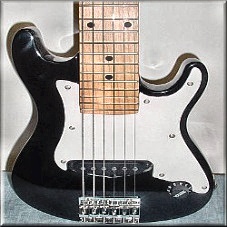}}\hspace{0.1cm} &
\fcolorbox{black}{black}{\includegraphics[width=1.4cm]{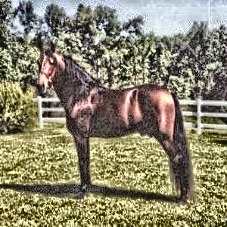}}\hspace{0.1cm} &
\fcolorbox{black}{black}{\includegraphics[width=1.4cm]{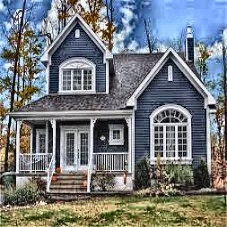}}\hspace{0.1cm} \\

\multicolumn{7}{c}{Convolutional} \\

\rotatebox{90}{Convolve}\hspace{0.1cm} &
\fcolorbox{black}{black}{\includegraphics[width=1.4cm]{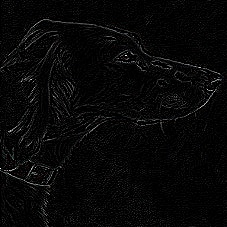}}\hspace{0.1cm} &
\fcolorbox{black}{black}{\includegraphics[width=1.4cm]{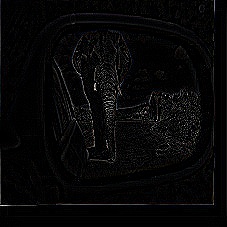}}\hspace{0.1cm} &
\fcolorbox{black}{black}{\includegraphics[width=1.4cm]{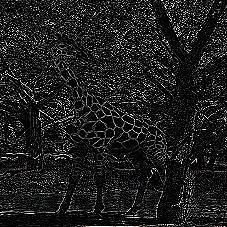}}\hspace{0.1cm} &
\fcolorbox{black}{black}{\includegraphics[width=1.4cm]{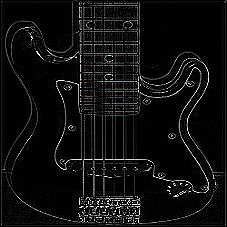}}\hspace{0.1cm} &
\fcolorbox{black}{black}{\includegraphics[width=1.4cm]{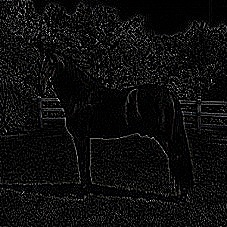}}\hspace{0.1cm} &
\fcolorbox{black}{black}{\includegraphics[width=1.4cm]{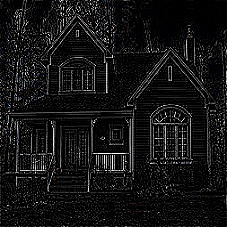}}\hspace{0.1cm} \\
\end{tabular}

}
\label{fig:aug5}
\end{subfigure}
\end{minipage}%
\begin{minipage}{.5\textwidth}
\centering
\begin{subfigure}{\textwidth}
\resizebox{\columnwidth}{!}{%
\setlength{\fboxsep}{0pt}%
\setlength{\fboxrule}{0.5pt}%
\scriptsize
\begin{tabular}{c@{}c@{}c@{}c@{}c@{}c@{}c@{}}
& Dog & Elephant & Giraffe & Guitar & Horse & House  \\[0.2cm] 

\multicolumn{7}{c}{Geometric cont.} \\

\rotatebox{90}{Elastic}\hspace{0.1cm} &
\fcolorbox{black}{black}{\includegraphics[width=1.4cm]{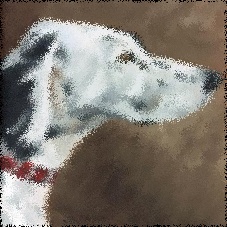}}\hspace{0.1cm} &
\fcolorbox{black}{black}{\includegraphics[width=1.4cm]{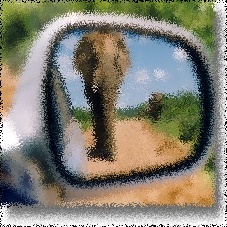}}\hspace{0.1cm} &
\fcolorbox{black}{black}{\includegraphics[width=1.4cm]{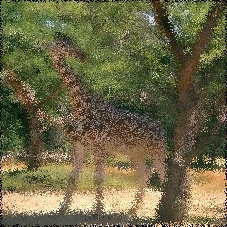}}\hspace{0.1cm} &
\fcolorbox{black}{black}{\includegraphics[width=1.4cm]{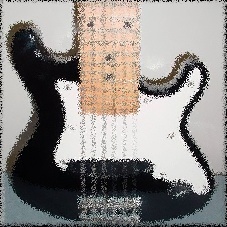}}\hspace{0.1cm} &
\fcolorbox{black}{black}{\includegraphics[width=1.4cm]{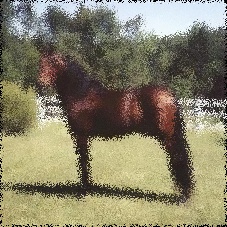}}\hspace{0.1cm} &
\fcolorbox{black}{black}{\includegraphics[width=1.4cm]{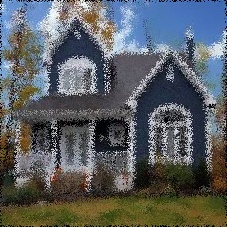}}\hspace{0.1cm} \\
\rotatebox{90}{Warping}\hspace{0.1cm} &
\fcolorbox{black}{black}{\includegraphics[width=1.4cm]{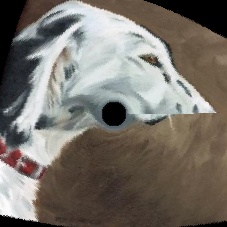}}\hspace{0.1cm} &
\fcolorbox{black}{black}{\includegraphics[width=1.4cm]{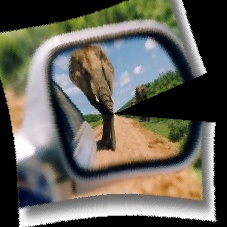}}\hspace{0.1cm} &
\fcolorbox{black}{black}{\includegraphics[width=1.4cm]{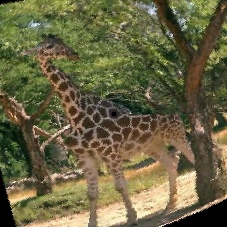}}\hspace{0.1cm} &
\fcolorbox{black}{black}{\includegraphics[width=1.4cm]{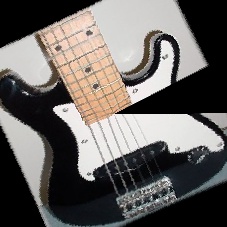}}\hspace{0.1cm} &
\fcolorbox{black}{black}{\includegraphics[width=1.4cm]{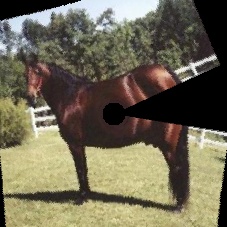}}\hspace{0.1cm} &
\fcolorbox{black}{black}{\includegraphics[width=1.4cm]{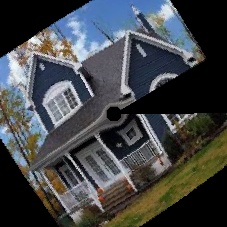}}\hspace{0.1cm} \\
\end{tabular}

}

\label{fig:aug6}
\end{subfigure}
\end{minipage}

\caption{Examples of augmentations used from each augmentation category.}
\label{fig:aug_combined3}
\vspace{-10pt}
\end{figure*}

\paragraph{Extra augmentations:} Examples from the PACS dataset of all 76 extra augmentations are shown in Figures \ref{fig:aug_combined1}-\ref{fig:aug_combined3}.

\clearpage

\end{document}